\documentclass[final]{cvpr}

\usepackage{times}
\usepackage{epsfig}
\usepackage{graphicx}
\usepackage{amsmath}
\usepackage{amssymb}
\usepackage{multirow}
\usepackage{colortbl}
\usepackage{booktabs}
\usepackage{overpic}

\usepackage{caption}
\usepackage{subcaption}
\usepackage{bbold}

\usepackage[pagebackref=true,breaklinks=true,colorlinks,bookmarks=false]{hyperref}

\def\cvprPaperID{4598} 
\def\confYear{CVPR 2021}

\begin{document}

\title{Triple-cooperative Video Shadow Detection}

\author{Zhihao Chen$^{1 \ast}$, Liang Wan$^{1}$\thanks{Zhihao Chen and Liang Wan are the joint first authors of this work.}\ , Lei Zhu$^{2}$\thanks{Lei Zhu (lz437@cam.ac.uk) is the corresponding author of this work.},
Jia Shen$^{1}$, Huazhu Fu$^{3}$, Wennan Liu$^{4}$, Jing Qin$^{5}$\\
    $^1$ College of Intelligence and Computing, Tianjin University \\
	$^2$ Department of Applied Mathematics and Theoretical Physics, University of Cambridge \\
	$^3$ Inception Institute of Artificial Intelligence, UAE\\
	$^4$ Academy of Medical Engineering and Translational Medicine, Tianjin University \\
	$^5$ The Hong Kong Polytechnic University
}


\maketitle

\thispagestyle{empty}


\begin{abstract}
Shadow detection in a single image has received significant research interests in recent years. However, much fewer works have been explored in shadow detection over dynamic scenes. The bottleneck is the lack of a well-established dataset with high-quality annotations for video shadow detection. 
In this work, we collect a new video shadow detection  dataset (ViSha), which contains $120$ videos with $11,685$ frames, covering 60 object categories, varying lengths, and different motion/lighting conditions. All the frames are annotated with a high-quality pixel-level shadow mask. To the best of our knowledge, this is the first learning-oriented dataset for video shadow detection.
Furthermore, we develop a new baseline model, named triple-cooperative video shadow detection network (TVSD-Net). It utilizes triple parallel networks in a cooperative manner to learn discriminative representations at intra-video and inter-video levels. Within the network, a dual gated co-attention module is proposed to constrain features from neighboring frames in the same video, while an auxiliary similarity loss is introduced to mine semantic information between different videos. 
Finally, we conduct a comprehensive study on ViSha, evaluating 12 state-of-the-art models (including single image shadow detectors, video object segmentation, and saliency detection methods). Experiments demonstrate that our model outperforms SOTA competitors.

\end{abstract}

\section{Introduction}
\label{sec::introduction}

As a common phenomenon in our daily life, shadows in natural images provide hints for extracting scene geometry~\cite{okabe2009attached,karsch2011rendering}, light direction~\cite{lalonde2009estimating}, and camera location and its parameters~\cite{junejo2008estimating}.
Shadows can also benefit diverse image understanding tasks, e.g., image segmentation~\cite{ecins2014shadow}, object detection~\cite{cucchiara2003detecting}, and object tracking~\cite{nadimi2004physical}.
The last decade has witnessed a growing interest in image shadow detection. Many methods have been developed by examining color and illumination priors~\cite{finlayson2006removal,finlayson2009entropy}, by developing data-driven approaches with hand-crafted features~\cite{huang2011characterizes,lalonde2010detecting,zhu2010learning},
or by learning discriminative features from a convolutional neural network (CNN)~\cite{khan2014automatic,vicente2016large,nguyen2017shadow,Hu_2018_CVPR,le2018a+,zhu2018bidirectional,hu2019direction,zheng2019distraction}.

However, in striking contrast with the flourishing development of image shadow detection, much fewer works have been explored in shadow detection over dynamic scenes. On the other hand, we also notice that video processing has become an urgent topic in recent years, and a lot of methods were proposed for video salient object detection~\cite{le2017deeply, wang2017video, fan2019shifting} and video object segmentation~\cite{lu2019see,oh2019video}. 
What makes video shadow detection lag far behind these video processing tasks? Compared with shadow detection of a single image, video shadow detection (VSD) needs to utilize temporal information to identify shadow pixels of each video frame. Although there exist multiple datasets for image shadow detection, video salient object detection, and video object segmentation, such standard widespread benchmark (with a sufficient number of video clips, covering diverse content) is missing for video shadow detection. What's more, CNN-based methods have not been exploited for this problem due to the lack of such a dataset.

\begin{figure*}
\centering
\includegraphics[width=0.92\linewidth]{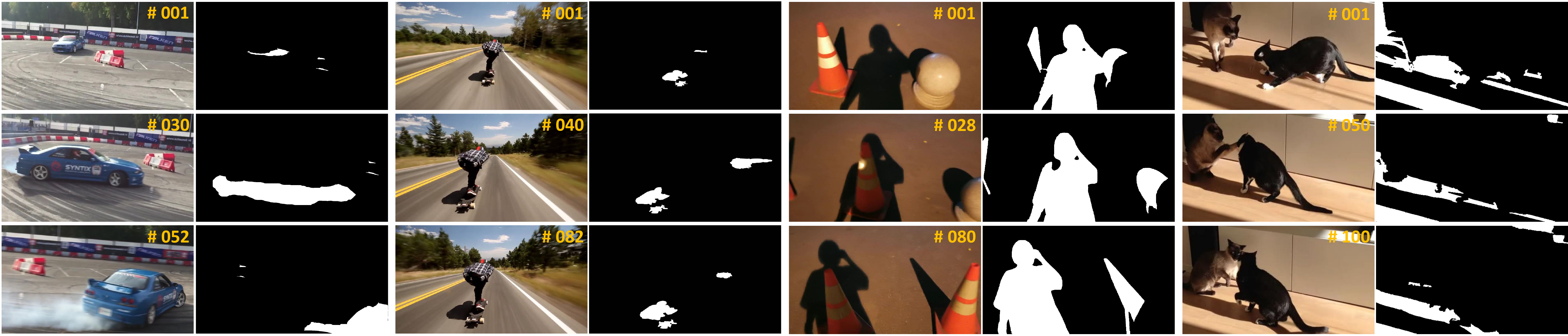}
\vspace{-1.5mm}
\caption{The examples of proposed Video Shadow Detection (\textit{\textbf{ViSha}}) dataset, with pixel-level shadow annotations.}
\vspace{-3mm}
\label{fig:dataset_profile}  
\end{figure*}

In this work, \textbf{we first collect a new video shadow detection (ViSha) dataset.} It contains 120 videos with 11,685 image frames and 390 seconds duration, covering shadows of 7 object classes and 60 object categories, various motion/lighting conditions, and different instance numbers. All the video frames are carefully annotated with a high-quality pixel-level shadow mask. To the best of our knowledge, this is the first learning-oriented dataset for video shadow detection, which could facilitate the community to explore further in this field.  
Second, \textbf{we develop a new baseline model, a triple-cooperative video shadow detection network (TVSD-Net), for this task.}
Instead of just exploiting temporal information within one video clip as most current video object detection networks did, we propose to learn at both intra-video and inter-video levels then model their correlation features. 
Our TVSD-Net utilizes triple parallel networks in a cooperative manner. 
To be specific, we take two neighboring frames from the same video and one image frame from another video as inputs. 
Then a dual gated co-attention (DGC) module is devised to learn a global intra-video correlation on the two frames of the same video, and a triple-cooperative (T) module encodes the inter-video property, which promotes the similarity between the same-video frames and suppresses the similarities between different-video frames.
Finally, \textbf{we present a comprehensive evaluation of 12 state-of-the-art models on our ViSha dataset}, making it the most complete VSD benchmark. Results show that our model significantly outperforms existing methods, including single image shadow detectors~\cite{chen2020multi, zhu2018bidirectional, zheng2019distraction}, single saliency detectors~\cite{houqibin2018DSS, deng2018r3net}, semantic segmentation method~\cite{lin2017feature, Zhao_2017_CVPR}, video object segmentation~\cite{lu2019see}, and video saliency detection methods~\cite{li2019motion, song2018pyramid}. 
In summary, our work forms the first learning-oriented VSD benchmark, thereby providing a new view to video object detection from a shadow perspective. Our  dataset and code have been released at \url{https://github.com/eraserNut/ViSha}.
\section{Related Works}
\label{sec:relatedworks}

\noindent\textbf{Single-image Shadow detection.}
Existing shadow detection works mainly focus on detect shadow pixels from a single input image. Deep learning-based methods have achieved dominated results. 
Please refer to~\cite{chen2020multi,wang2019densely,hosseinzadeh2018fast} for a review on traditional shadow detectors based on hand-crafted features. 
The first shadow detection CNN~\cite{khan2014automatic} identified shadow pixels by building a seven-layer CNN to extract deep features from superpixels, and then employed a conditional random field (CRF) to further smooth shadow detection results.
Vicente et al.~\cite{vicente2016large} trained a patch-based CNN with an image-level shadow prior.
Nguyen~\cite{nguyen2017shadow} introduced a generative adversarial network with a conditional generator to generate a shadow mask.
Hu et al.~\cite{Hu_2018_CVPR} learned spatial context features in a direction-aware manner, while Zhu et al.~\cite{zhu2018bidirectional} utilized two series of recurrent attention residual (RAR) modules to aggregate context information at different CNN layers for shadow detection. 
Zheng et al.~\cite{zheng2019distraction} learned distraction-aware features to explicitly
predict false positives and false negatives for robust shadow detection.
Rather than relying on only annotated shadow data, Chen et al.~\cite{chen2020multi} explored the complementary information of shadow region detection, shadow boundary detection, and shadow count detection and embedded the multi-task learning into a semi-supervised learning framework to fuse unlabeled data for helping shadow detection.
%

\definecolor{Rowcolor}{RGB}{230,230,230}
\begin{table*}[!t]
\begin{center}
  \caption{Statistics of the proposed ViSha dataset. See Section~\ref{subsec:statistics} for details.} 
  \label{table:statistics}
  \resizebox{\textwidth}{!}
{%
\begin{tabular}{c||cc||cc||cccc||cc||cccc}
    \toprule[1pt]
    \multirow{2}*{\large{\textbf{\textsl{ViSha}}}} & \multicolumn{2}{c||}{Shadow Motion} &
    \multicolumn{2}{c||}{Camera Motion} & \multicolumn{4}{c||}{\# Shadow Instances} & \multicolumn{2}{c||}{Shadow Type} &
    \multicolumn{4}{c}{Scene Type}
    \\
     & \cellcolor{Rowcolor} Stable & \cellcolor{Rowcolor}Moving & \cellcolor{Rowcolor}Stable & \cellcolor{Rowcolor}Moving & \cellcolor{Rowcolor}1 & \cellcolor{Rowcolor}2 & \cellcolor{Rowcolor}3 & \cellcolor{Rowcolor}$\geq$4 & \cellcolor{Rowcolor}Hard & \cellcolor{Rowcolor}Soft & \cellcolor{Rowcolor}Day & \cellcolor{Rowcolor}Night& \cellcolor{Rowcolor}Indoor & \cellcolor{Rowcolor}Outdoor  \\
     \hline
     \hline
    \# Training  & 10 & 40 & 12 & 38 & 8 & 4 & 6 & 22 & 36 & 12 & 33 & 17& 18 & 32 \\
    \hline
    \# Testing & 10 & 60 & 20 & 50 & 17 & 15 & 9 & 29 & 52 & 18   & 53 & 17& 17 & 53 \\
     \hline
     \hline
    \# Total & 20 & 100 & 32 & 88 & 25 & 19 & 15 & 51 & 88 & 30 &  86 & 34& 35 & 85 \\
    \bottomrule[1pt]
\end{tabular}
}
  \end{center} 
  \vskip -15pt
\end{table*}

\vspace{2mm}
\noindent\textbf{Video shadow detection} aims to detect the shadow regions from each frame of a video. Existing video shadow detection methods almost relied on hand-crafted features and were developed one decade ago. For instance, Nadimi et al.~\cite{nadimi04} leveraged a spatio-temporal
albedo test and dichromatic reflection model. Jr et al.~\cite{Jr05} detected the moving shadow in videos by employing improved background subtraction techniques. 
Benedek et al.~\cite{Benedek08} combined the color and microstructural features to detect the shadow in surveillance Videos. Note that these methods work well on high-quality scenarios (\eg, stable lighting, single shadow, moving objects) due to the limited generalization capability of these hand-craft features. In addition, no large-scale datasets are publicly available for fairly evaluating different video shadow detection methods. 
In order to exploit the capability of CNN-based methods for VSD tasks, it is desirable to collect a large-scale VSD dataset and develop a CNN model to provide a complement evaluation.

\vspace{2mm}
\noindent\textbf{Video object segmentation} automatically detects primary foreground objects from their background in all frames of a video.
It can be roughly categorized into unsupervised video object segmentation (UVOS) and semi-supervised video object segmentation (SVOS) (please refer to~\cite{lu2019see,oh2019video} for a detailed review). 
Compared with the counterpart for image object segmentation, VOS exploits the temporal information across frames.
Lu et al.~\cite{lu2019see} formulated a co-attention siamese network (COSNet) to model UVOS from a global perspective via a co-attention mechanism.
Oh et al.~\cite{oh2019video} leveraged memory networks and learned to read relevant information from all past frames with object masks for resolving SVOS.
However, current CNN-based VOS mainly learned appearance or motion
representations in intra-video, while ignoring the valuable discriminative inter-video representations across different videos.

\begin{table}[t]
\centering
  \caption{Video sources of our ViSha dataset.} 
  \vspace{-1mm}
  \label{table:collection}
  \resizebox{0.48\textwidth}{!}{%
    \begin{tabular}{c||c|c|c|c|c|c}
        \toprule[1pt]
        Source & OTB~\cite{OTB2015} & VOT~\cite{votpami} & LaSOT~\cite{lasot} & TC-128~\cite{TC128} & NfS~\cite{NfS} & Self
        \\
        \hline
        \# Videos  & 11 & 7 & 18 & 16 & 9 & 59 \\
        \bottomrule[1pt]
    \end{tabular}
    } 
    \vskip -5pt
\end{table}

\paragraph{Video saliency detection} identifies most distinctive objects for each video frame~\cite{cheng2014global,wang2015saliency,aytekin2017spatiotemporal,8704996,8466906,SODsurvey}.
Recently, many CNN-based video saliency detection methods achieved dominant results.
The first CNN attempt was a fully convolutional network proposed by Wang et al.~\cite{wang2017video}.
Le et al.~\cite{le2017deeply} adopted 3D filters to combine spatial and temporal information in a spatio-temporal CRF framework, while Li et al.~\cite{li2018flow}  presented optical flow guided recurrent neural network.
Song et al.~\cite{song2018pyramid} passed concatenated spatial features at multiple scales into an extended deeper bidirectional ConvLSTM to obtain spatio-temporal information.
Fan et al.~\cite{fan2019shifting} collected a video saliency detection dataset and developed a saliency-shift-aware convLSTM module to extract both spatial and temporal information.
Similar to the task of video object detection, video saliency detection methods usually focus on extracting spatial-temporal features from multiple frames of a video, which also ignores the intra-video discriminative property of salient objects.


\begin{figure}[!t]
\centering
\includegraphics[scale=.235]{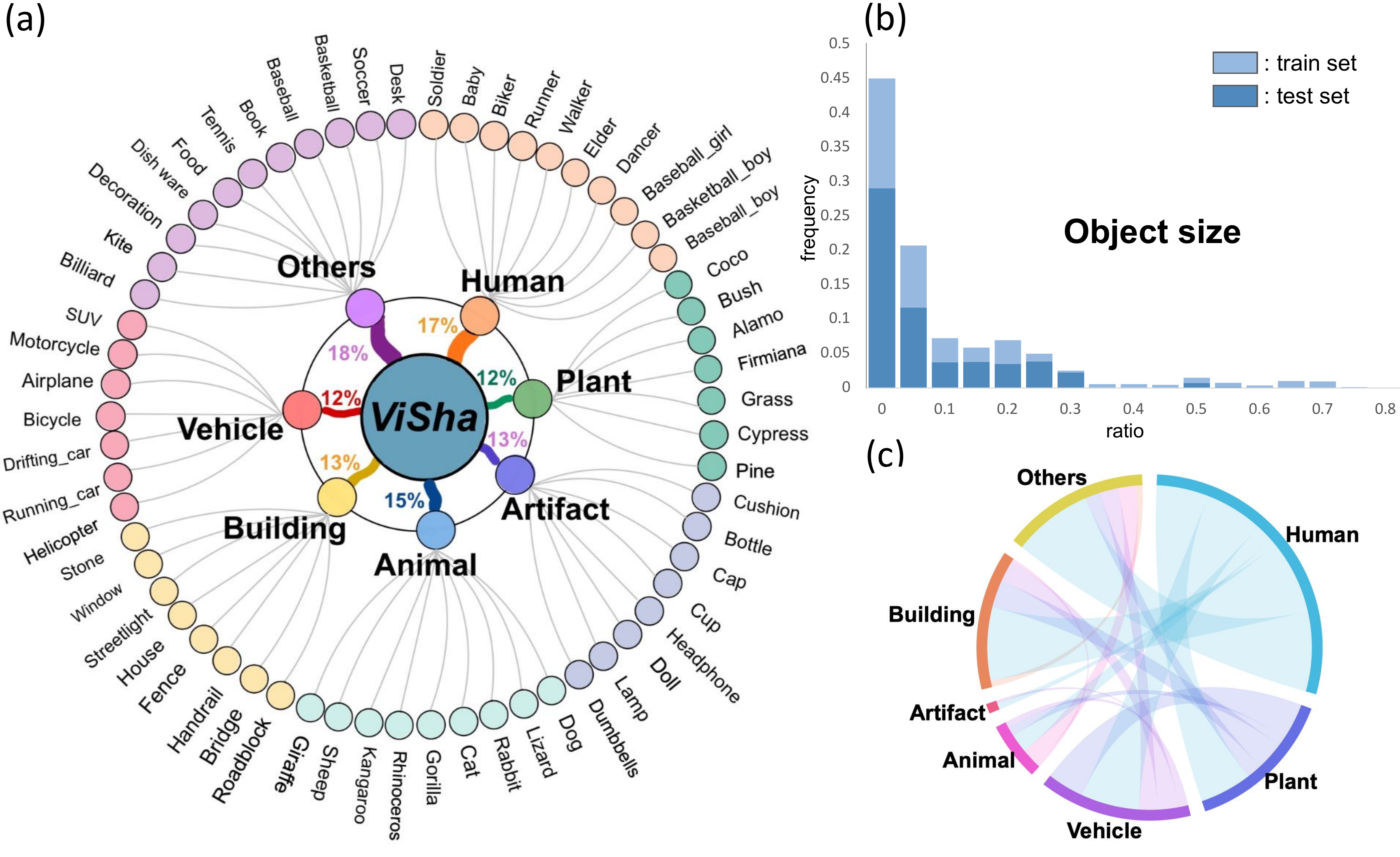}
\vspace{-1mm}
\caption{Statistics of the proposed ViSha. (a) Shadow categories. (b) Ratio distribution of the shadows. (c) Mutual dependencies among shadow categories in (a).}
\label{fig:dataset_relation}  
\vskip -10pt
\end{figure}

\section{ViSha: Video Shadow Detection Dataset}
\label{sec:dataset}

We introduce \textbf{ViSha}, a new dataset for video shadow detection. Our dataset includes 120 videos with diverse content, varying length, and object-level annotations. Some example video frames can be found in Figure~\ref{fig:dataset_profile}.    
We will show details of ViSha from the following key aspects.

\subsection{Data Collection} \label{subsec:collection}
In order to provide a solid basis for video shadow detection, we think that the dataset should (1) cover diverse realistic-scenes, and (2) contain sufficient challenging cases.  
As shown in Table~\ref{table:collection}, more than half videos are from 5 widely-used video tracking benchmarks (\ie, OTB~\cite{OTB2015}, VOT~\cite{votpami}, LaSOT~\cite{lasot}, TC-128~\cite{TC128}, and NfS~\cite{NfS}). 
Note that these video tracking datasets are not originally designed for shadow detection, and hence there are limited  videos with shadows, which are all included in our dataset.  
The remaining 59 videos are self-captured with different handheld cameras, over different scenes, at varying times. 
We then manually trim the videos to make sure that each frame has at least one shadow area, and remove dark-screen transitions. 
The frame rate is adjusted to 30 fps for all video sequences. For instance, the videos from NfS~\cite{NfS} have a high-speed frame rate of 240, for which we make sampling at every 8 frames.
Eventually, our video shadow detection dataset (ViSha) contains 120 video sequences, with a totally of 11,685 frames and 390 seconds duration. The longest video contains 103 frames and the shortest contains 11 frames.


\begin{figure*}[!t]
\centering
\includegraphics[scale=.35]{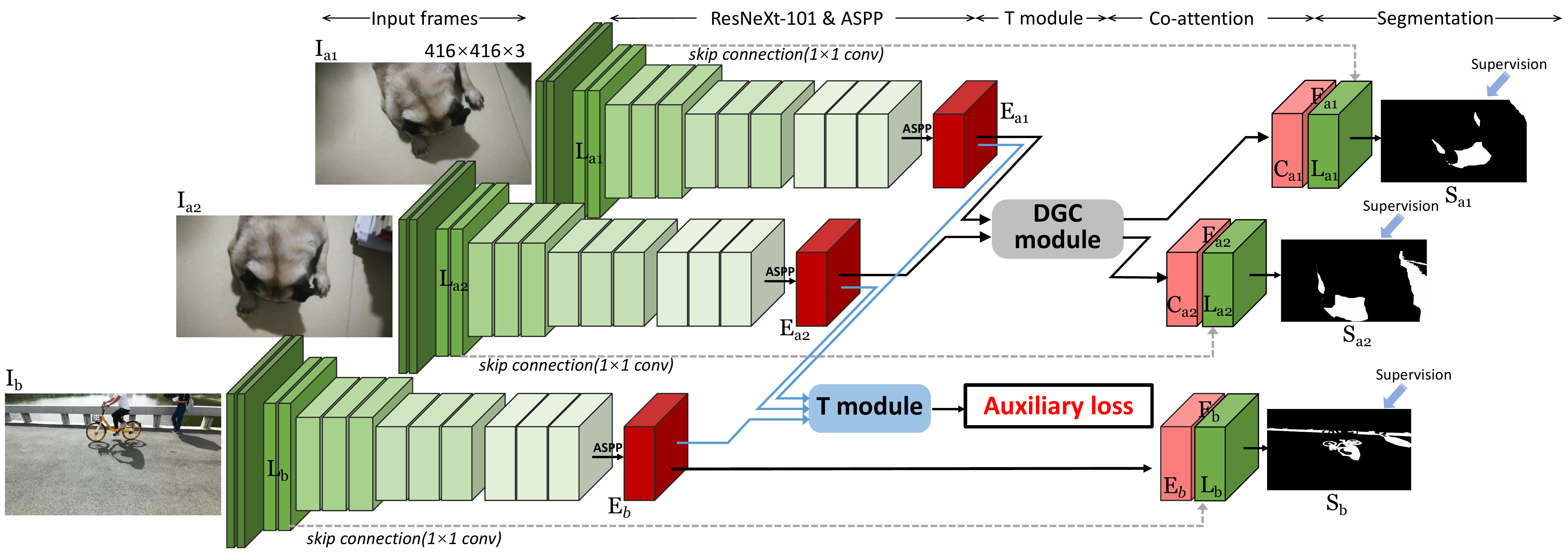} 
\vskip -5pt
\caption{The schematic illustration of our proposed TVSD-Net.  DGC module denotes dual-gated co-attention module and T module denotes triple-cooperative module. See Section~\ref{sec:fullArchitecture} for details.}
\label{fig:architecture} 
\vspace{-3mm}
\end{figure*}

\subsection{Dataset Annotation and Split} \label{subsec:annotation}
For each video frame, we provide pixel-accurate, manually created segmentation in the form of a binary mask. 
In realistic scenarios, shadows can be distorted, ambiguous , and hard to identify (see examples in Figure~\ref{fig:dataset_profile}).  
Eight human annotators are pretrained and instructed to carefully annotate all the shadows by tracing shadow boundaries. 
Then, two viewers are assigned to inspect and validate the labeled shadows. 
In the annotation process, we notice that two cases deserve  special attention. 
First, the soft shadow is usually subjected to unclear boundaries. Considering the temporal consistency between adjacent frames, we demand that the labeling of soft shadows should be consistent across frames. 
Second, the back-light parts of objects often appear in dark colors, yet they do not form shadows, for which we treat them as non-shadow areas. 
%

To provide guidelines for future works, we randomly split the  dataset into training and testing sets with a ratio of 5:7. Table~\ref{table:statistics} shows the statistics for the dataset. We can see that both the training set and testing set have sufficient diversity. It is also worth noting that we allocate more video sequences for testing sets because small testing sets may lead to model over-fitting.

\begin{figure*}[!t]
\centering
\includegraphics[scale=.52]{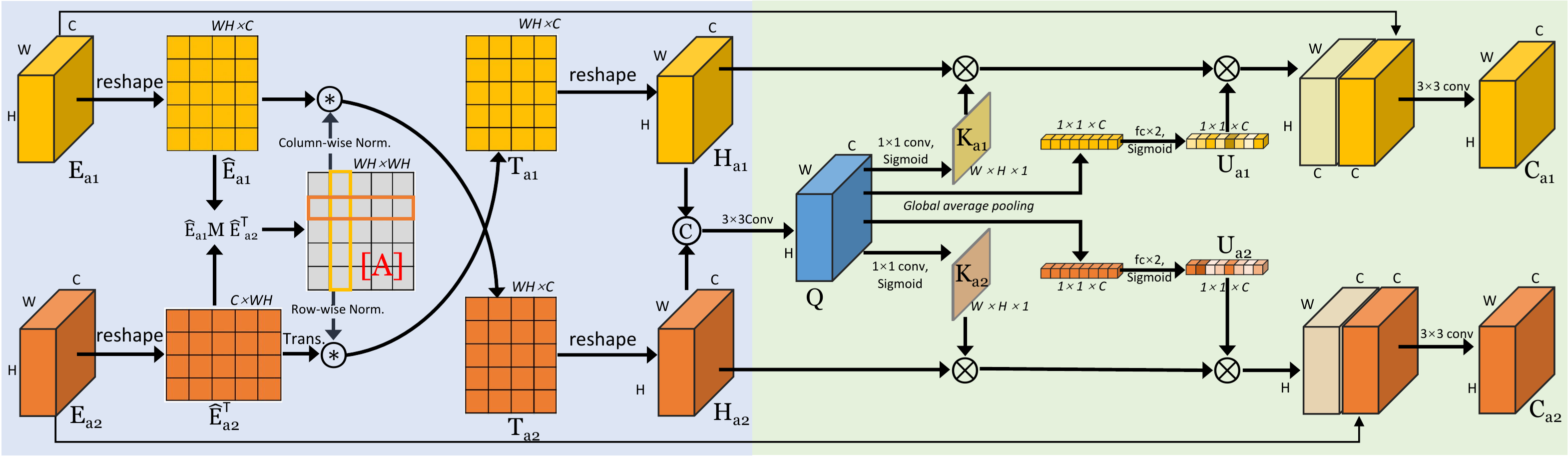}
\vspace{-3.5mm}
\caption{The schematic illustration of Dual Gated Co-attention (DGC) Module. Blue and green parts represent the collaboration attention (co-attention) module and dual-gated module, respectively. See Section~\ref{sec:DGC} for details.}
\label{fig:co-attention}
\vspace{-2.5mm}
\end{figure*}

\subsection{Dataset Features and Statistics} \label{subsec:statistics}

\vspace{2mm}
\noindent\textbf{Sufficient Shadow Diversity.}
The diversity of shadow refers to the diversity of shadow sources, i.e. objects which cast the corresponding shadows. In ViSha, the shadow source is composed of seven main categories: Human, Plant, Vehicle, Animal, Artifact, Building, and Others. Figure~\ref{fig:dataset_relation} (a)\&(c) show these categories (7 main classes with 60 sub-classes) and their mutual dependencies, respectively. Figure~\ref{fig:dataset_relation} (b) shows the shadow ratio distribution in ViSha. The shadow ratio is deﬁned as the proportion of the number of shadow pixels to that of the entire image. Besides, as shown in Table~\ref{table:statistics}, there are 95 videos having more than one shadow instance, making this dataset challenging for the video shadow detection task.

\vspace{2mm}
\noindent\textbf{Motion of Camera and Objects.}
As a video dataset, ViSha contains ample motion diversity for objects and cameras (summarized in Table~\ref{table:statistics}). In the viewpoint of shadow motion, 20 videos have the object shadows stay relatively static to the background, while the other 100 videos witness shadows suffer from fast-moving or distortion. Similarly, from the aspect of camera motion, there are 32 videos in which the camera is fixed and the shadow changes in the scene are relatively stable (\eg, surveillance cameras), while in the rest 88 videos, the shadows exhibit drastic changes and/or motion blur lead by camera shaking or movement.

\vspace{2mm}
\noindent\textbf{Various Lighting Conditions.}
Different lighting conditions can lead to hard shadow, which has an obvious boundary, or soft shadow, which has a rather blurry or unclear boundary. Hard shadow is usually produced when the scene contains only one single strong light source (\eg, sunshine). Soft shadow is usually caused under lighting condition with multiple light sources. 88 videos in Visha contain the hard shadows while 30 videos contain the soft shadows.

\vspace{2mm}
\noindent\textbf{Richness of the Scene.}
As we all know, data-driven models are subjected to the domain shift. For example, if all videos in the training set of ViSha are taken in the daytime, the trained models can hardly handle the shadows in night scenes. The same phenomenon also applies to the cases of indoor and outdoor scenes. In order to avoid such a problem, we build ViSha with 86 daytime videos and 34 night videos. Furthermore, ViSha contains 35 indoor videos and 85 ourdoor videos. More examples of ViSha can be found in the supplementary material. 
\section{Proposed Method} \label{sec:method}



\subsection{Overview of Our Network} \label{sec:fullArchitecture}


Figure~\ref{fig:architecture} shows the schematic illustration of our triple-cooperative video shadow detection network (TVSD-Net). The intuition behind our network is to leverage discriminative feature information at both intra-video and inter-video levels. That is, for neighboring frames from the same video, their features shall be similar; while frames from different videos will have features to be distinguishable. 

Our TVSD-Net takes three shadow images as inputs. The first two images (denoted as $\{\mathbf{I}_{a1}$, $\mathbf{I}_{a2}\}$) are from the same video, while the third image $\mathbf{I}_b$ is randomly selected from another video. 
We devise three branches to pass each input image into a feature embedding module to extract three high-level semantic features, which are denoted as $\{\mathbf{E}_{a1}, \mathbf{E}_{a2}, \mathbf{E}_b\} \in\mathbb{R}^{26\times26\times256}$.
The feature embedding module consists of a feature extraction backbone (ResNeXt-101) and an atrous spatial pyramid pooling (ASPP) module, which has a $1$$\times$$1$ point-wise convolution, three $3$$\times$$3$ convolutions with dilation rates of 12, 24, and 36 respectively, and a global average pooling layer.
We empirically replace the last CNN layers of ResNeXt-101 with the dilation convolution (dilation rate of 2) and set the first convolutional stride to 1 to balance the spatial resolution of features and GPU memory size.

To learn global intra-video features, we devise a co-attention mechanism to emphasize the coherent information in $\mathbf{E}_{a1}, \mathbf{E}_{a2}$ from the same video (DGC module; see Figure~\ref{fig:co-attention}). 
The refined features are denoted as $\mathbf{C}_{a1}$ and $\mathbf{C}_{a2}$, respectively.
Note that deep CNN layers are able to capture highly semantic features tending to describe global attributes of shadow regions, while  shallow CNN layers are responsible for extracting subtly fine features to represent delicate structures.
We concatenate the refined high-level feature $\mathbf{C}_{a1}$ with a low-level feature map $\mathbf{L}_{a1}$ from the feature extraction backbone via a ship connection in the first network branch and then apply $3$$\times$$3$ and $1$$\times$$1$ convolutional layers on the concatenated features to generate a shadow detection result $\mathbf{S}_{a1}$.
Similarly, the second branch concatenates $\mathbf{C}_{a2}$ with a low-level feature map $\mathbf{L}_{a2}$ of the feature extraction backbone to generate another shadow detection result $S_{a2}$.


In the third branch, without any co-attention module, we directly concatenate high-level features $\mathbf{E}_b$ with a low-level feature map $\mathbf{L}_{b}$, and predict one more shadow detection result $\mathbf{S}_{b}$.
What's more, we devise a triple-cooperative module (T module; see Section~\ref{sec:auxiliaryTask}) to learn inter-video features in helping shadow detection.
The auxiliary loss adopted in T module makes $\mathbf{E}_{a1}$ and $\mathbf{E}_{a2}$ from two frames of the same video similar, while $\mathbf{E}_{b}$ from another video should be dissimilar to them.

\vspace{2mm}
\noindent\textbf{Loss Function.}
To better handle the scale variance of shadows, 
we fuse the binary cross entropy (BCE) loss function with a lov\'asz-hinge loss~\cite{berman2018lovasz} function to compute the shadow detection loss ($\mathcal{L}_{seg}$) of all three inputs $I_{a_1}$, $I_{a_2}$, and $I_{b}$:
\vspace{-2mm}
\begin{equation}\label{Equ:seg_loss}
    \mathcal{L}_{seg} = \mathcal{L}_{a1} + \mathcal{L}_{a2} + \mathcal{L}_{b} , \vspace{-2mm}
\end{equation} 
where
\vspace{-2mm}
\begin{equation}\label{Eq:total_consistency_loss_2}
  \begin{aligned}
    & \mathcal{L}_{a1} = \Phi_{BCE}(\mathbf{S}_{a1}, \mathbf{G}_{a1}) + \Phi_{Hinge}(\mathbf{S}_{a1}, \mathbf{G}_{a1}) \ , \\
    & \mathcal{L}_{a2} = \Phi_{BCE}(\mathbf{S}_{a2}, \mathbf{G}_{a2}) + \Phi_{Hinge}(\mathbf{S}_{a2}, \mathbf{G}_{a2}) \ , \\
    & \mathcal{L}_{b} = \Phi_{BCE}(\mathbf{S}_{b}, \mathbf{G}_{b}) + \Phi_{Hinge}(\mathbf{S}_{b}, \mathbf{G}_{b}) \ .
  \end{aligned}
\end{equation}
Here, $\Phi_{BCE}(\cdot)$ and $\Phi_{Hinge}(\cdot)$ denote the BCE loss and the lov\'asz-hinge loss, respectively;
$\mathbf{S}_{a1}$/$\mathbf{S}_{a2}$/$\mathbf{S}_{b}$ and $\mathbf{G}_{a1}$/$\mathbf{G}_{a2}$/$\mathbf{G}_{b}$ are the predicted shadow detection map and the corresponding ground truth of $\mathbf{I}_{a1}$/$\mathbf{I}_{a2}$/$\mathbf{I}_{b}$;

Finally, we use a combination of the shadow detection segmentation loss $\mathcal{L}_{seg}$ and the devised auxiliary task loss $\mathcal{L}_{aux}$ (described in Section~\ref{sec:auxiliaryTask}) to train our whole network. The total loss of our network is given by:
\vspace{-2mm}
\begin{equation}\label{Equ:total_loss}
    \mathcal{L}_{total} = \mathcal{L}_{seg} + \beta \mathcal{L}_{aux},
\end{equation}
where $\beta$ is to control the weight of auxiliary loss, and we empirically set $\beta = 10$ in our experiments.

\subsection{Dual Gated Co-attention Module} \label{sec:DGC}


The dual gated co-attention module explicitly encodes intra-video correlations between a pair of frames in a video, via a co-attention mechanism and a dual-gated mechanism. This enables TVSD-Net to focus on frequently coherent regions, thus further helping to discover the shadow regions and produce reasonable VSD results.


\vspace{2mm}
\noindent\textbf{Co-attention Mechanism.}  
The blue region of Figure~\ref{fig:co-attention} shows the collaboration-attention (co-attention) module, which takes two features $\mathbf{E}_{a1}$ $\in \mathbb{R}^{W \times H \times C}$, $\mathbf{E}_{a2}$ as the inputs to compute their correlations. 
Inspired by \cite{lu2019see}, we first reshape $\mathbf{E}_{a1}$ to be a new feature map $\hat{\mathbf{E}}_{a1}$ $\in \mathbb{R}^{WH \times C}$ and reshape $\mathbf{E}_{a2}$ to be  $\hat{\mathbf{E}}_{a2}$ $\in \mathbb{R}^{WH \times C}$, and compute an affinity matrix $\mathbf{A} \in \mathbb{R}^{WH \times WH}$:
\vspace{-2mm}
\begin{equation}\label{Equ:affinity}
    \mathbf{A} = \hat{\mathbf{E}}_{a1}\mathbf{M} \hat{\mathbf{E}}_{a2}^{\top},
\end{equation}
where $\mathbf{M} \in \mathbb{R}^{C \times C}$ is a weight matrix. 
Intuitively, each element of $\mathbf{A}$ represents the similarity between each column of $\mathbf{E}_{a1}$ and each row of $\mathbf{E}_{a2}$.


From $\mathbf{A}$, we employ a $\mathtt{Softmax}$ function to column-wisely and row-wisely normalize $\mathbf{A}$ respectively, and multiply the resultant normalization features with $\mathbf{A}$ to compute two co-attention enhanced features $\mathbf{T}_{a1}$ and $\mathbf{T}_{a2}$:
\begin{equation}\label{Equ:transform}
\begin{aligned}
    \mathbf{T}_{a1} &= \mathtt{Softmax}(\mathbf{A})*\hat{\mathbf{E}}_{a2}  \in \mathbb{R}^{C \times WH} \ , \\
    \mathbf{T}_{a2} &= \mathtt{Softmax}(\mathbf{A}^{\top})*\hat{\mathbf{E}}_{a1}   \in \mathbb{R}^{C \times WH} \ ,
\end{aligned}
\end{equation}
Then, we reshape $\mathbf{T}_{a1}$ to be $\mathbf{H}_{a1} \in \mathbb{R}^{C \times W \times H}$ and reshape $\mathbf{T}_{a2}$ to be $\mathbf{H}_{a2} \in \mathbb{R}^{C \times W \times H}$.
By this way, we intuitively transform the $a1$ features ($\mathbf{E}_{a1}$) to a fake $a2$ features ($\mathbf{H}_{a2}$). Compared to the original $\mathbf{E}_{a2}$, the fake one ($\mathbf{H}_{a2}$) encodes more temporal information.


\vspace{2mm}
\noindent\textbf{Dual-gated Mechanism.}
Since there may exist potential appearance variations (\eg, occlusion, out-of-view) between two neighboring frames,
using co-attention module enhances the coherent features, yet may also introduce some noises from adjacent frames. 
Hence, it is better to weight co-attention enhanced features from two input frames, instead of treating the learned co-attention information equally.
To achieve this goal, we propose a dual-gated mechanism to obtain co-attention confidences. 

Unlike the self-gated mechanism~\cite{lu2019see}, we learn the co-attention confidences by leveraging $\mathbf{H}_{a1}$ and $\mathbf{H}_{a2}$ together. Our dual-gated mechanism consists of a spatial gated operation and a channel gated operation.

Specifically, we fuse $\mathbf{H}_{a1}$ and $\mathbf{H}_{a2}$ by applying a $3$ $\times$ $3$ convolution on the concatenation of $\mathbf{H}_{a1}$ and $\mathbf{H}_{a2}$ to compute a fused feature map $\mathbf{Q}$:
\vspace{-2mm}
\begin{equation}\label{Equ:feature-fusion}
    \mathbf{Q} = \mathtt{Conv}(\mathtt{Concat}(\mathbf{H}_{a1}, \mathbf{H}_{a2})) \ .
\end{equation}
Then, two spatial gated maps ($\{\mathbf{K}_{a1}, \mathbf{K}_{a2}\} \in \mathbb{R}^{W \times H \times 1}$) are computed by utilizing a $\mathtt{Sigmoid}$ function and a $1$ $\times$ $1$ convolution on $\mathbf{Q}$:
\vspace{-2mm}
\begin{equation}\label{Equ:spatial_gate}
    \begin{aligned}
    \mathbf{K}_{a1} & = \mathtt{Sigmoid}(\mathtt{Conv}(\mathbf{Q})) \ , \\
    \mathbf{K}_{a2} & = \mathtt{Sigmoid}(\mathtt{Conv}(\mathbf{Q})) \ .
    \end{aligned}
\end{equation}
\noindent Moreover, we generate two channel-wise gated maps $\mathbf{U}_{a1}$ and $\mathbf{U}_{a2}$:
\vspace{-2mm}
\begin{equation}\label{Equ:channel_gate}
    \begin{aligned}
    \mathbf{U}_{a1} &= \mathtt{Sigmoid}(\mathtt{fc}(\mathtt{GAP}(\mathbf{Q})))  \ , \\
    \mathbf{U}_{a2} &= \mathtt{Sigmoid}(\mathtt{fc}(\mathtt{GAP}(\mathbf{Q})))  \ . \\
    \end{aligned}
\end{equation}

Once obtaining spatial and channel gated maps, we multiply the spatial gated map with the co-attention enhanced features $\{\mathbf{H}_{a1}, \mathbf{H}_{a2}\}$, and then multiply the resultant features with the channel gate map to produce gated features $\{\mathbf{D}_{a1}, \mathbf{D}_{a2}\}$.
We then apply a $3$ $\times$ $3$ convolution on the concatenation of $\mathbf{D}_{a1}$/$\mathbf{D}_{a2}$ and $\mathbf{E}_{a1}$ ($\mathbf{E}_{a2}$) to produce output features of the dual gated co-attention module, i.e., $\mathbf{C}_{a1}$ and $\mathbf{C}_{a2}$. 
The definitions of $\mathbf{C}_{a1}$ and $\mathbf{C}_{a2}$ are given by:
\begin{equation}\label{Equ:final_co_attention}
\begin{aligned}
    \mathbf{C}_{a1} &= \mathtt{Conv}(\mathtt{Cancat}(\mathbf{E}_{a1}, \mathbf{H}_{a1} \otimes \mathbf{K}_{a1} \otimes \mathbf{U}_{a1})) \ , \\
    \mathbf{C}_{a2} &= \mathtt{Conv}(\mathtt{Cancat}(\mathbf{E}_{a2}, \mathbf{H}_{a2} \otimes \mathbf{K}_{a2} \otimes \mathbf{U}_{a2})) \ ,
\end{aligned}
\end{equation}
where $\otimes$ denotes element-wise product.

\subsection{Triple-cooperative Module} \label{sec:auxiliaryTask}
\begin{figure}[!t]
\centering
\includegraphics[scale=.5]{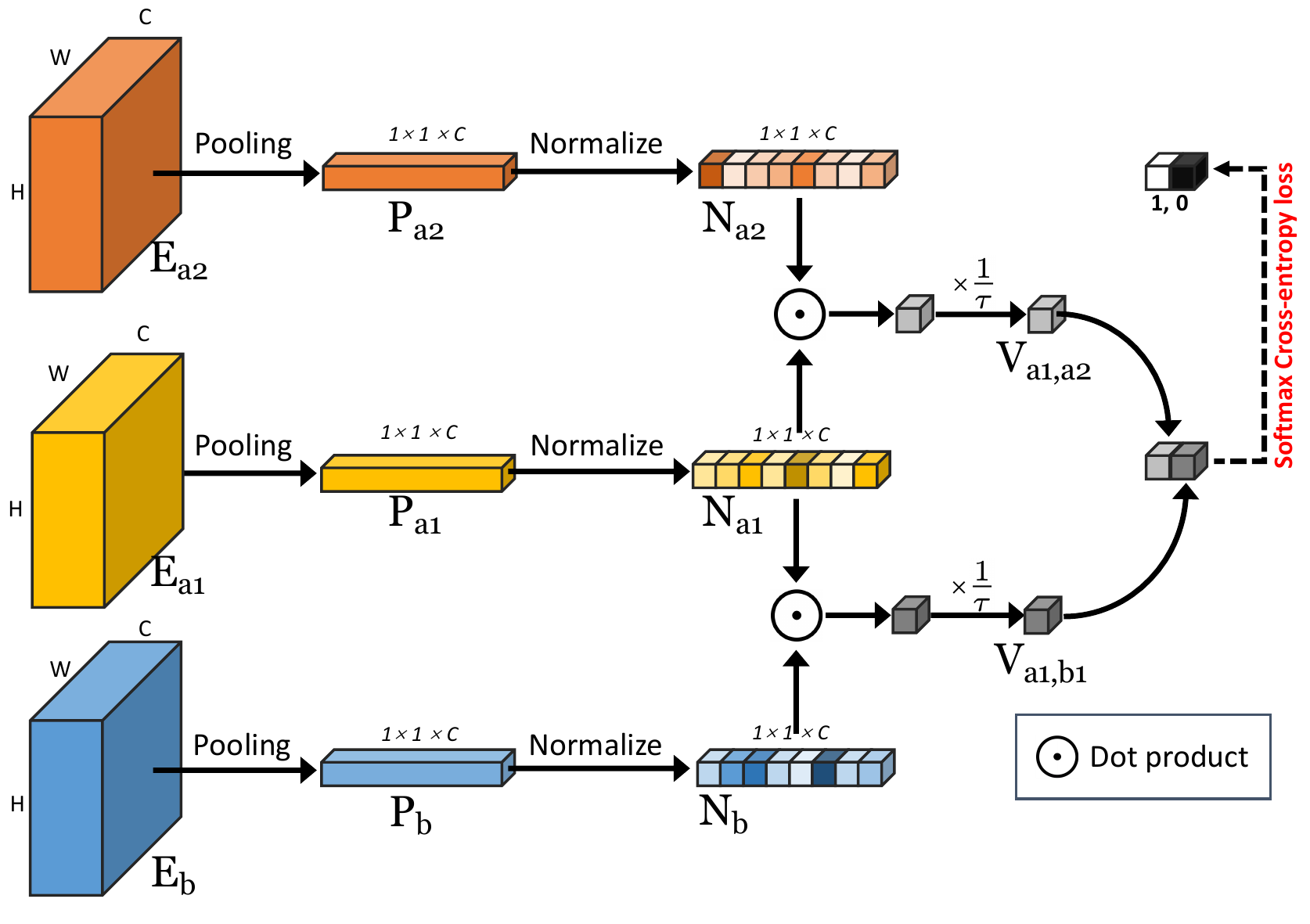}
\vskip -5pt
\caption{The schematic illustration of our triple-cooperative (T) module; See Section~\ref{sec:auxiliaryTask} for details.}
\label{fig:Auxiliary}
\vspace{-2.5mm}
\end{figure}
Given high-level features $\{\mathbf{E}_{a1}, \mathbf{E}_{a2}, \mathbf{E}_b\}$ of the three input images $\{\mathbf{I}_{a1}, \mathbf{I}_{a2}, \mathbf{I}_b\}$, we devise a triple-cooperative module (T-module) to make features from the same video similar and features from different videos dissimilar.
Figure~\ref{fig:Auxiliary} shows the schematic illustration of T-module, which computes an auxiliary loss based on $\{\mathbf{E}_{a1}, \mathbf{E}_{a2}, \mathbf{E}_b\}$. 
Intuitively, we expect the similarity between two frames from the same video to be close to 1, while the similarity between two frames from different videos approaches to 0.

To be specific, we apply three global average pooling operations on $\mathbf{E}_{a1}$, $\mathbf{E}_{a2}$, and $\mathbf{E}_b$ to obtain three features $\{\mathbf{P}_{a1}, \mathbf{P}_{a2}, \mathbf{P}_b\} \in \mathbb{R}^{1\times1\times256}$, which are then normalized as $\{\mathbf{N}_{a1}, \mathbf{N}_{a2}, \mathbf{N}_b\}$: 
\begin{equation}\label{Equ:normalize}
\begin{aligned}
    \mathbf{N}_{a1} &= \frac{\mathbf{P}_{a1}}{
    \mathtt{max}({\small \left \| \mathbf{P}_{a1} \right \|_2}, \epsilon)} \ ,  \;
    \mathbf{N}_{a2} = \frac{\mathbf{P}_{a2}}{
    \mathtt{max}({\small \left \| \mathbf{P}_{a2} \right \|_2}, \epsilon)} \ , \\
    \mathbf{N}_{b} &= \frac{\mathbf{P}_{b}}{
    \mathtt{max}({\small \left \| \mathbf{P}_{b} \right \|_2}, \epsilon)} \ , 
\end{aligned}
\end{equation}
where $\epsilon$ is a small positive number and it is set as $\epsilon$ $=$ $1e^{-12}$ to avoid division by zero.

After that, we compute the similarity $V_{a1, a2}$ of $\mathbf{N}_{a1}$ and $\mathbf{N}_{a2}$ from the same video by computing the dot produce of $\mathbf{N}_{a1}$ and $\mathbf{N}_{a2}$, and also the similarity $V_{a1, b}$ of $\mathbf{N}_{a1}$ and $\mathbf{N}_{b}$ from different videos, given by:
\vspace{-2mm}
\begin{equation}\label{Equ:similarity} 
       V_{a1, a2}=  \mathbf{N}_{a1} \cdot \mathbf{N}_{a2}, \; \text{and,} \;
       V_{a1, b} = \mathbf{N}_{a1} \cdot \mathbf{N}_{b}  . 
\end{equation}
Then, we multiple two similarities with a temperature  $\frac{1}{\tau}$~\cite{hinton2015} and concatenated them to form a two-element vector. The rest is to compare whether this two-element vector is close to the target distribution of (1, 0). Here, after applying a softmax function on the two-element vector for normalization, we compute a cross-entropy loss between the two vectors as the auxiliary loss of our T module.

In summary, the definition of our auxiliary loss $\mathcal{L}_{aux}$ is given by: 
\vspace{-2mm}
\begin{equation}\label{Equ:similarityLoss}  
    \mathcal{L}_{aux} = -\textrm{log}\frac{\textrm{exp}( V_{a1, a2}/ \tau)}{\textrm{exp}(V_{a1, a2} / \tau) + \textrm{exp}( V_{a1, b} / \tau)},
\end{equation}
where $\tau$$=$$0.7$ is a temperature constant to control degree of two similarities. The sum is over one positive and one negative samples.
It is clear that the auxiliary loss makes the similarity $V_{a1, a2}$ from same video to be 1 while making the similarity of $V_{a1, b}$ from different videos to be 0.
Hence, $\mathcal{L}_{aux}$ tends to have a small score when $\mathbf{N}_{a1}$ is similar to $\mathbf{N}_{a2}$ from the same video, and dissimilar to $\mathbf{N}_b$ from a different video.

\subsection{Implementation Details} \label{sec:implementationDetails}
We implement our TVSD-Net using PyTorch. 
Adam optimizer is employed to train the network with mixed precision training~\cite{micikevicius2017mixed} on a NVIDIA GTX 2080Ti.
We initialize the feature extraction backbone via a pre-trained ResNeXt-101~\cite{xie2017aggregated} on ImageNet while other layers are trained from scratch.
The weight decay, batch size, epoch number are set as $0.0005$, $5$, and $12$, respectively.
We set the initial learning rate as $0.0005$ for scratch layers and $0.00005$ for pretrained layers, and then use the cosine decay with a warm-up period to adjust the learning rate.
TVSD-Net requires about 0.06s to process an image of $416$ $\times$ $416$.


In the testing phase, we take the shadow detection result $\mathbf{S}_{a1}$ in the first branch as the output of the  TVSD-Net.
Given an input video, to obtain the shadow detection result of each frame (we call it target frame), we follow ~\cite{lu2019see} to empirically select the subsequent five frames of the target frame, and then pass the target frame as $\mathbf{I}_{a1}$, and each of fives frames as $\mathbf{I}_{a2}$ to the TVSD-Net.
By doing so, we obtain five segmentation results and then average the five results as the final shadow detection result of the target frame.
\section{Experiments}
\label{sec:experiments}
\subsection{Experimental Settings}

\vspace{0.5mm}
\noindent\textbf{Evaluation Metrics.} 
We adopt four common evaluation metrics to quantitatively compare video shadow detection methods. 
They are Mean Absolute Error (\textbf{MAE})~\cite{houqibin2018DSS,zhu2018saliency} and F-measure ($\mathbf{F}_{\beta}$)~\cite{houqibin2018DSS,zhu2019aggregating}, Intersection over Union (\textbf{IoU})~\cite{zhao2018icnet}, and Balance Error Rate (\textbf{BER})~\cite{Hu_2018_CVPR}.
In general, a better video shadow detection method shall have smaller BER and MAE scores, and larger $\mathbf{F}_{\beta}$  and IoU scores.

\vspace{0.5mm}
\noindent \textbf{Comparative Methods.} Since there is no CNN-based method for video shadow detection, we make comparison against $12$ state-of-the-art methods for relevant tasks, including FPN~\cite{lin2017feature}, PSPNet~\cite{Zhao_2017_CVPR}, DSS~\cite{houqibin2018DSS}, R$^3$Net~\cite{deng2018r3net}, BDRAR~\cite{zhu2018bidirectional}, DSD~\cite{zheng2019distraction}, MTMT~\cite{chen2020multi}, PDBM~\cite{song2018pyramid}, COSNet~\cite{lu2019see}, MGA~\cite{li2019motion}, FEELVOS~\cite{Voigtlaender_2019_CVPR} and STM~\cite{oh2019video}.
Among them, FPN and PSPNet are developed for single-image semantic segmentation. DSS and R$^3$Net are dedicated for single-image saliency detection, while BDRAR, DSD, and MTMT are utilized for single-image shadow detection.
Lastly, PDBM, COSNet, MGA, FEELVOS and STM are for video saliency detection and object object segmentation.
We use their public codes, and re-train these methods on our training set for a fair comparison.

\begin{figure*}[t]
	\centering

	\ \\
	\vspace*{0.5mm}
    \begin{subfigure}{0.085\textwidth} 
		\includegraphics[width=\textwidth]{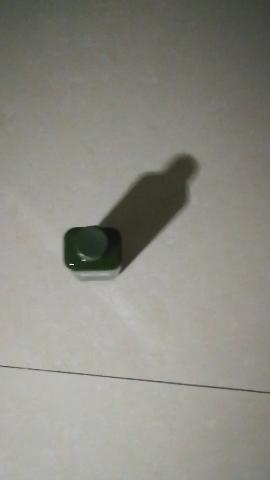}
	\end{subfigure}
	\begin{subfigure}{0.085\textwidth}
		\includegraphics[width=\textwidth]{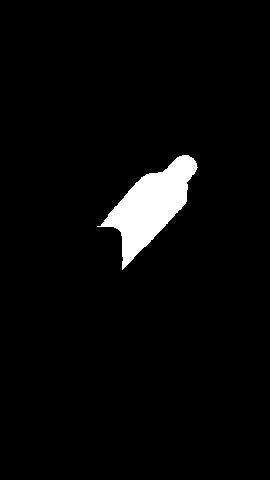}
	\end{subfigure}
	\begin{subfigure}{0.085\textwidth}
		\includegraphics[width=\textwidth]{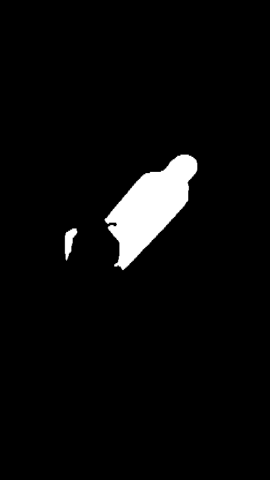}
	\end{subfigure}
	\begin{subfigure}{0.085\textwidth}
		\includegraphics[width=\textwidth]{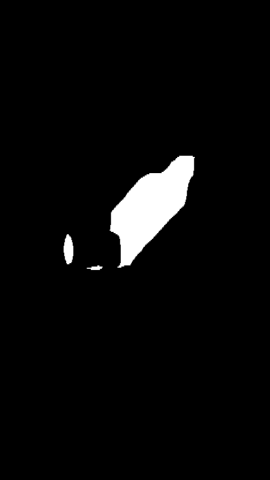}
	\end{subfigure}
	\begin{subfigure}{0.085\textwidth}
		\includegraphics[width=\textwidth]{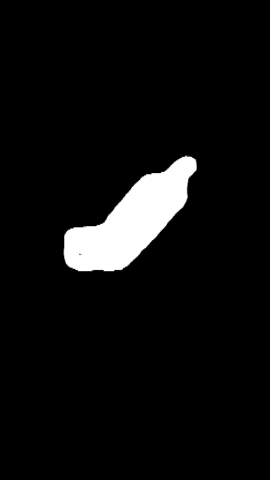}
	\end{subfigure}
	\begin{subfigure}{0.085\textwidth}
		\includegraphics[width=\textwidth]{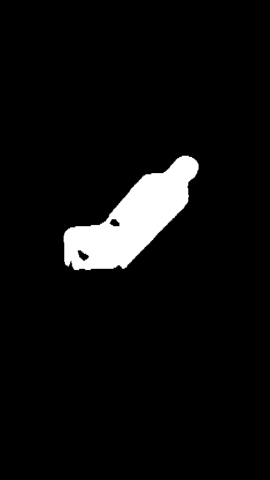}
	\end{subfigure}
	\begin{subfigure}{0.085\textwidth}
		\includegraphics[width=\textwidth]{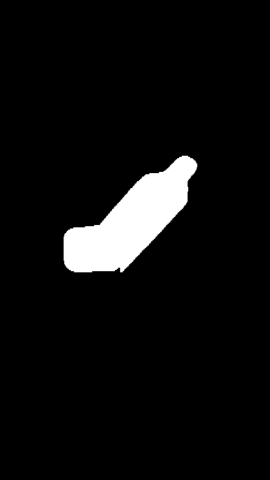}
	\end{subfigure}
	\begin{subfigure}{0.085\textwidth}
		\includegraphics[width=\textwidth]{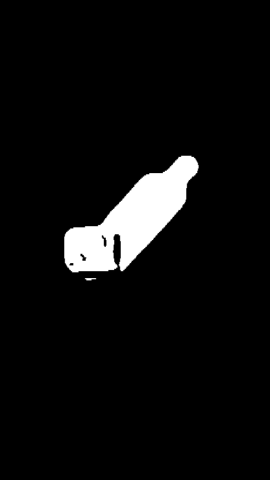}
	\end{subfigure}
	\begin{subfigure}{0.085\textwidth}
		\includegraphics[width=\textwidth]{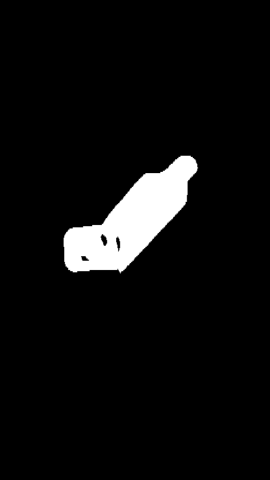}
	\end{subfigure}	
	\begin{subfigure}{0.085\textwidth}
		\includegraphics[width=\textwidth]{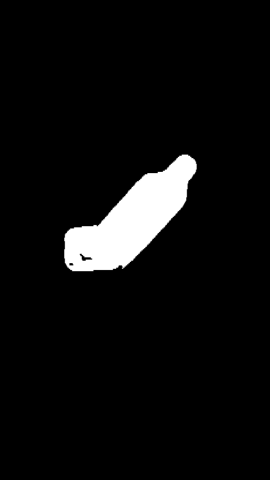}
	\end{subfigure}
	\begin{subfigure}{0.085\textwidth}
		\includegraphics[width=\textwidth]{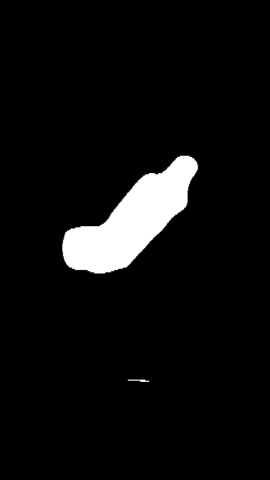}
	\end{subfigure}
	
	\ \\
	\vspace*{0.5mm}
    \begin{subfigure}{0.085\textwidth} 
		\includegraphics[width=\textwidth]{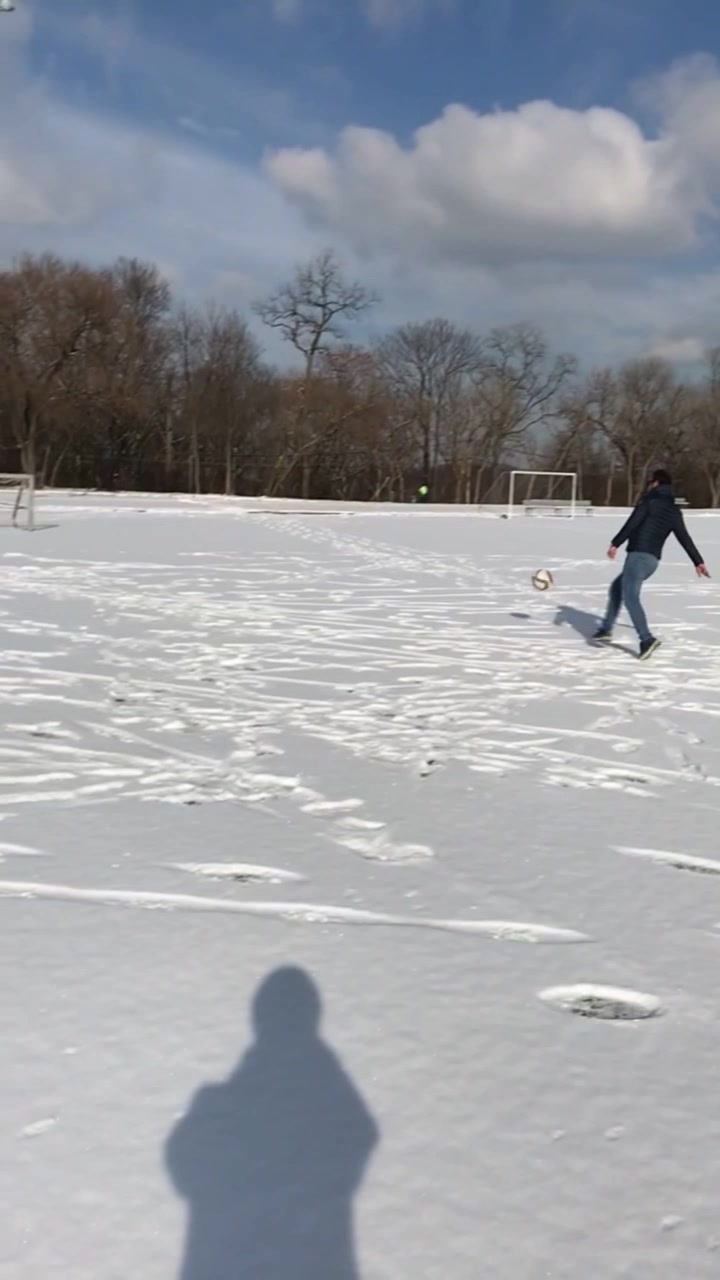}
	\end{subfigure}
	\begin{subfigure}{0.085\textwidth}
		\includegraphics[width=\textwidth]{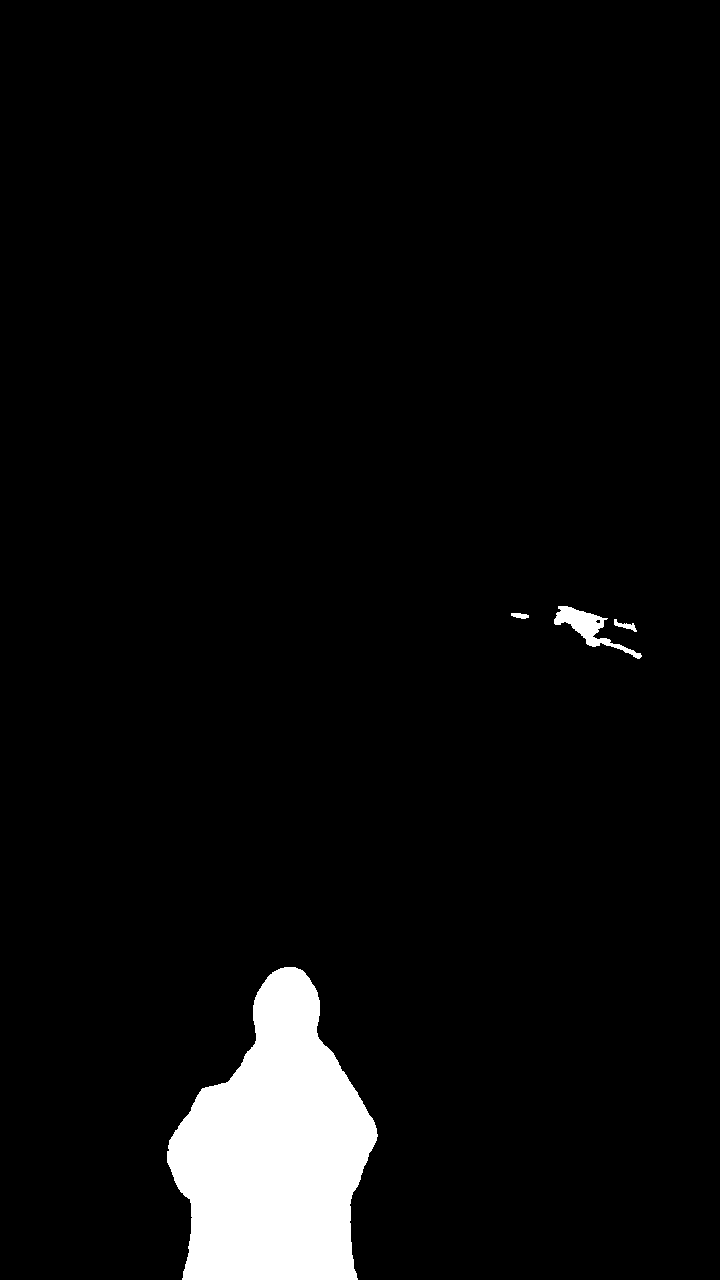}
	\end{subfigure}
	\begin{subfigure}{0.085\textwidth}
		\includegraphics[width=\textwidth]{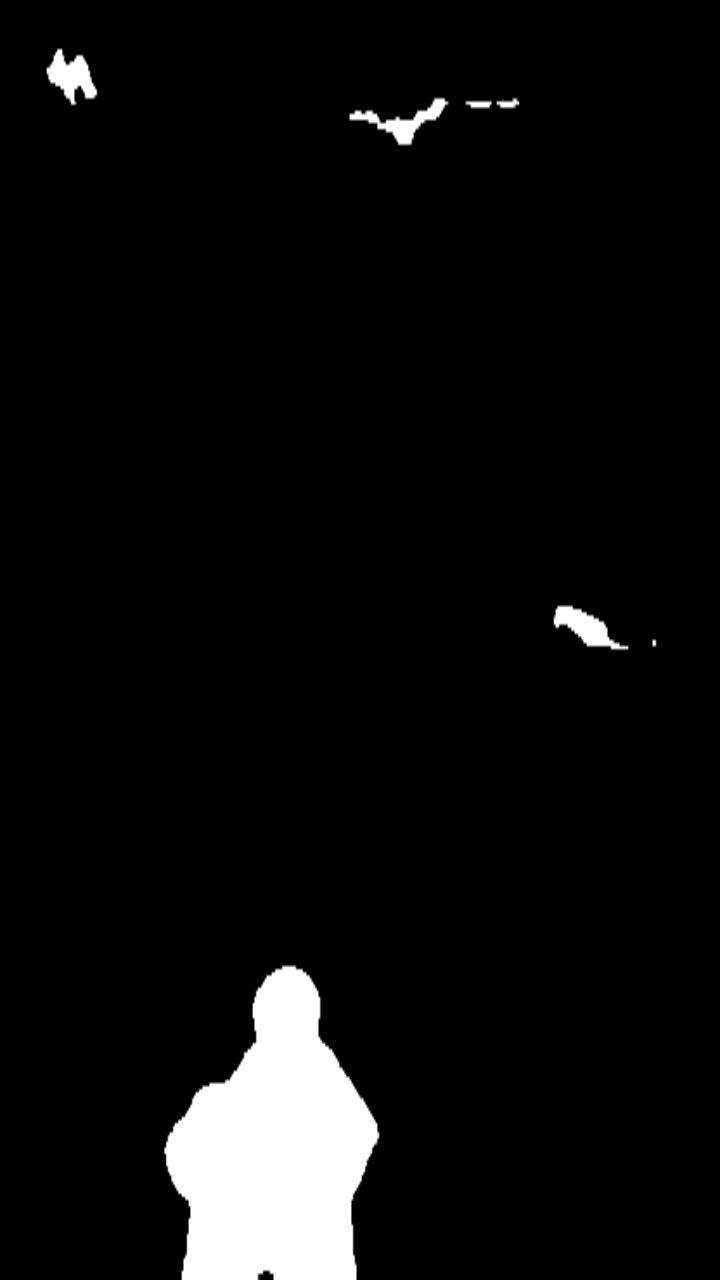}
	\end{subfigure}
	\begin{subfigure}{0.085\textwidth}
		\includegraphics[width=\textwidth]{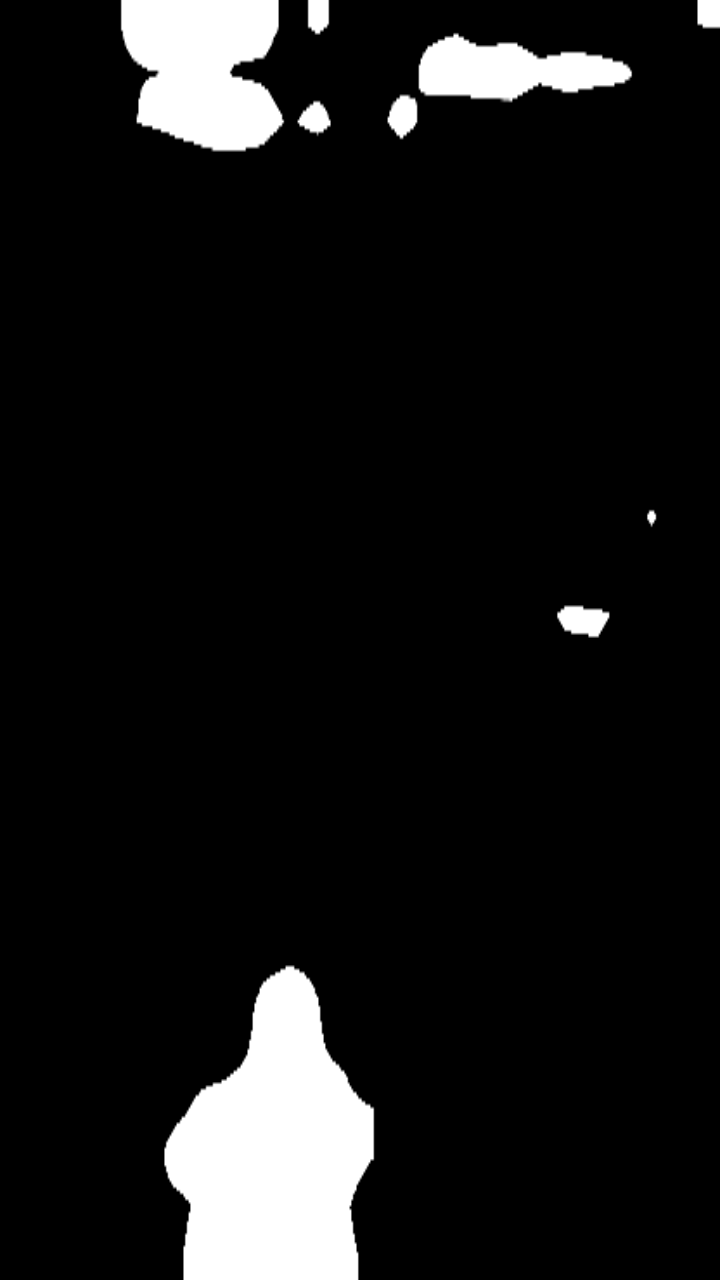}
	\end{subfigure}
	\begin{subfigure}{0.085\textwidth}
		\includegraphics[width=\textwidth]{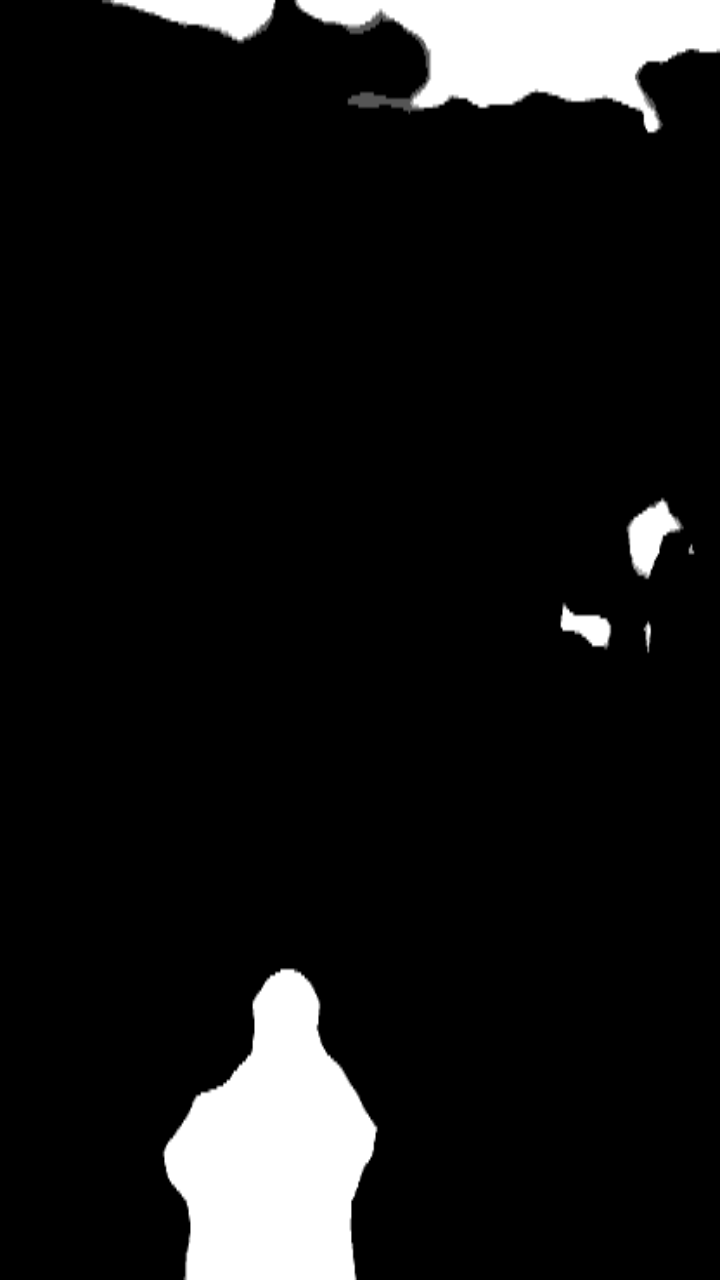}
	\end{subfigure}
	\begin{subfigure}{0.085\textwidth}
		\includegraphics[width=\textwidth]{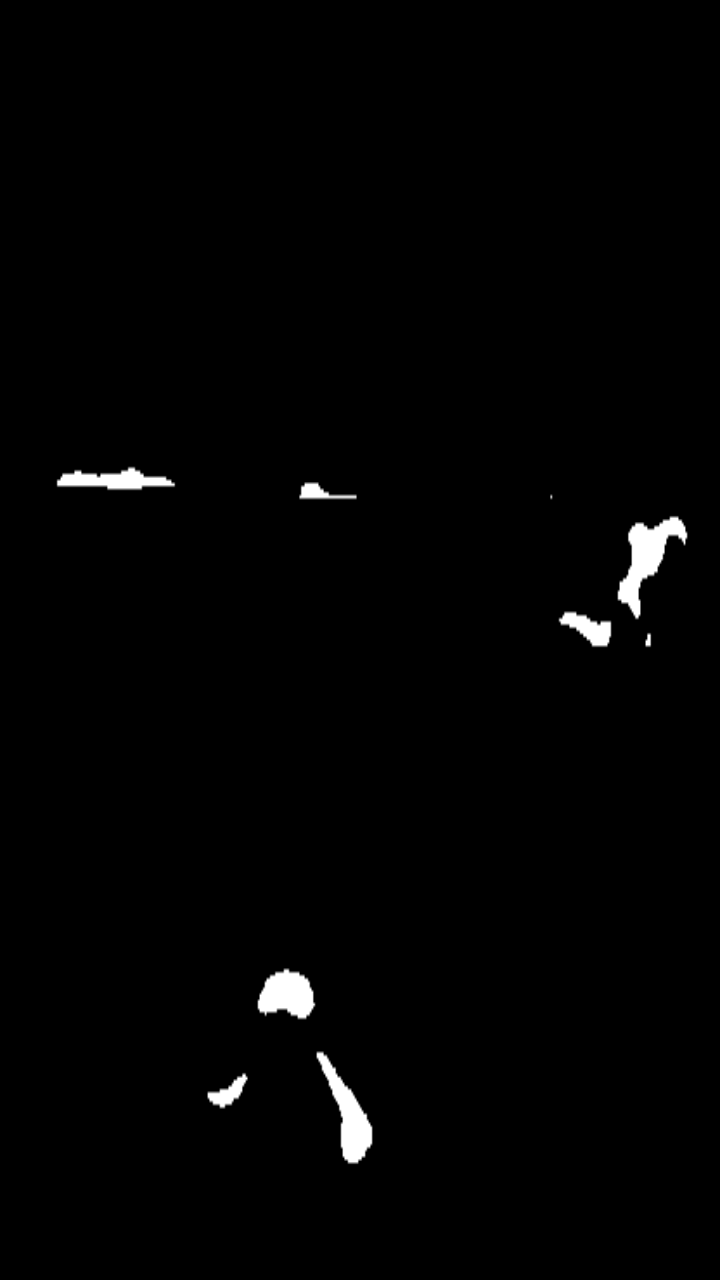}
	\end{subfigure}
	\begin{subfigure}{0.085\textwidth}
		\includegraphics[width=\textwidth]{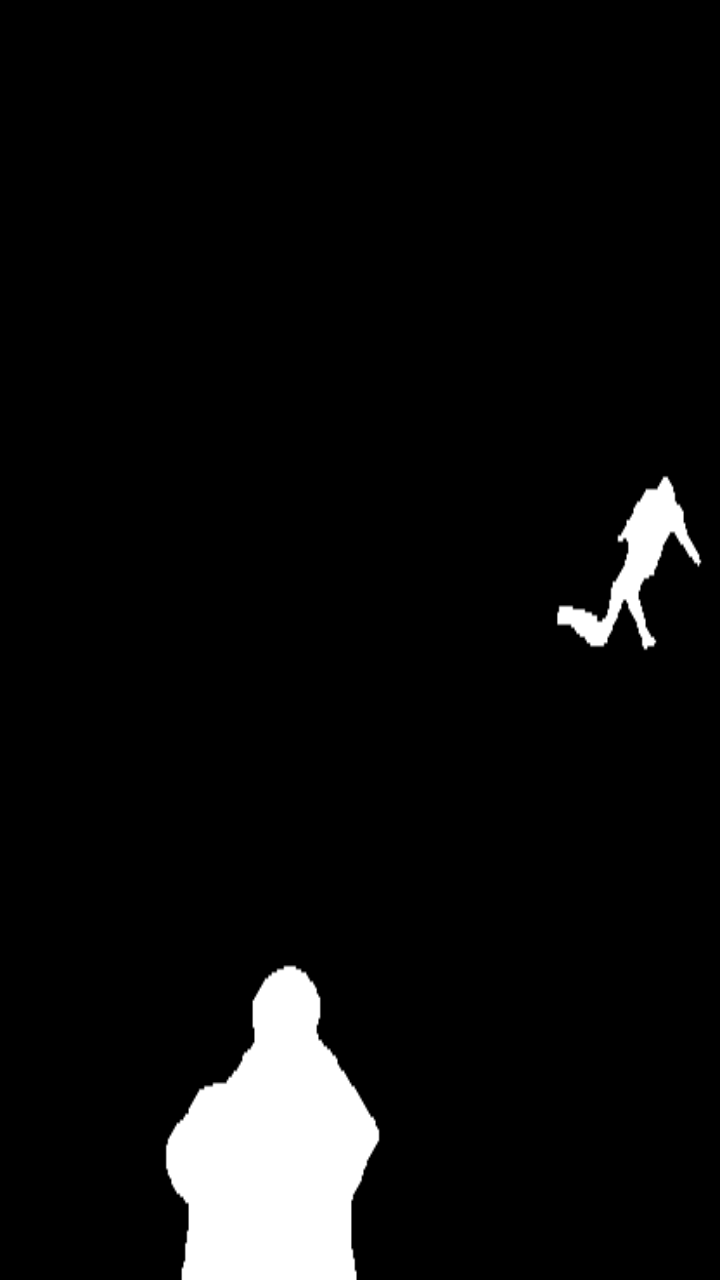}
	\end{subfigure}
	\begin{subfigure}{0.085\textwidth}
		\includegraphics[width=\textwidth]{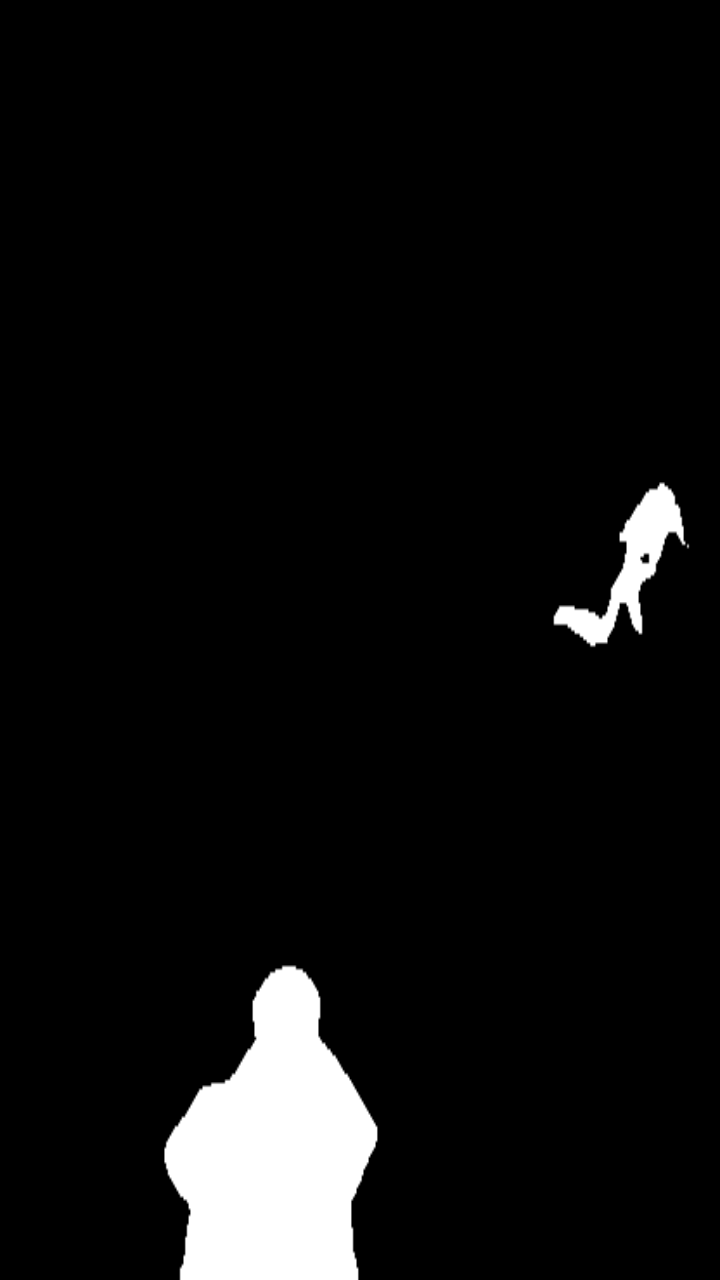}
	\end{subfigure}
	\begin{subfigure}{0.085\textwidth}
		\includegraphics[width=\textwidth]{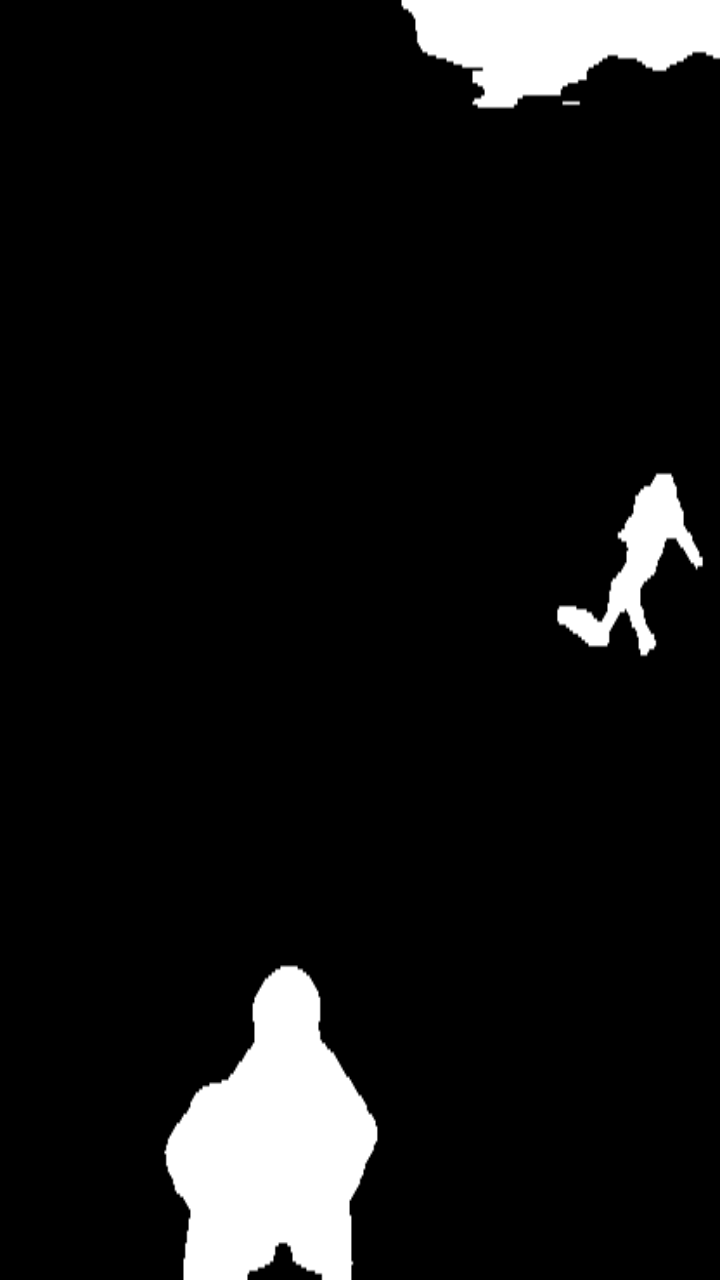}
	\end{subfigure}	
	\begin{subfigure}{0.085\textwidth}
		\includegraphics[width=\textwidth]{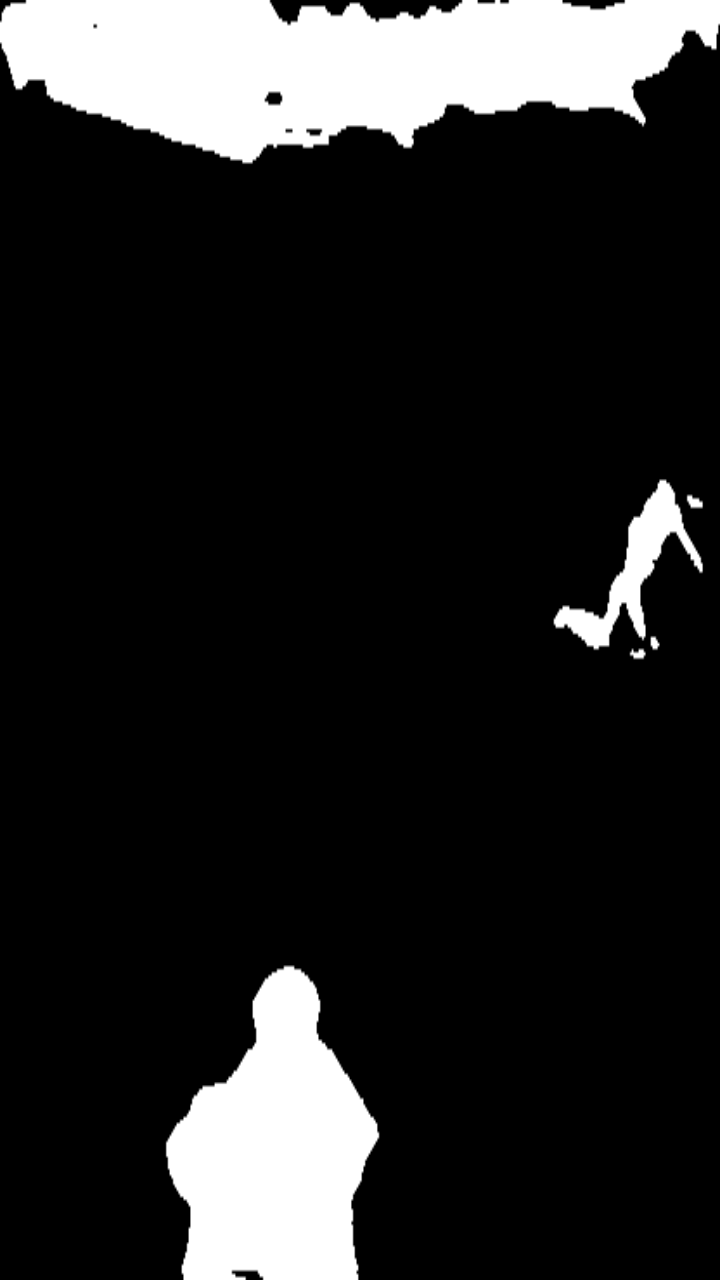}
	\end{subfigure}
	\begin{subfigure}{0.085\textwidth}
		\includegraphics[width=\textwidth]{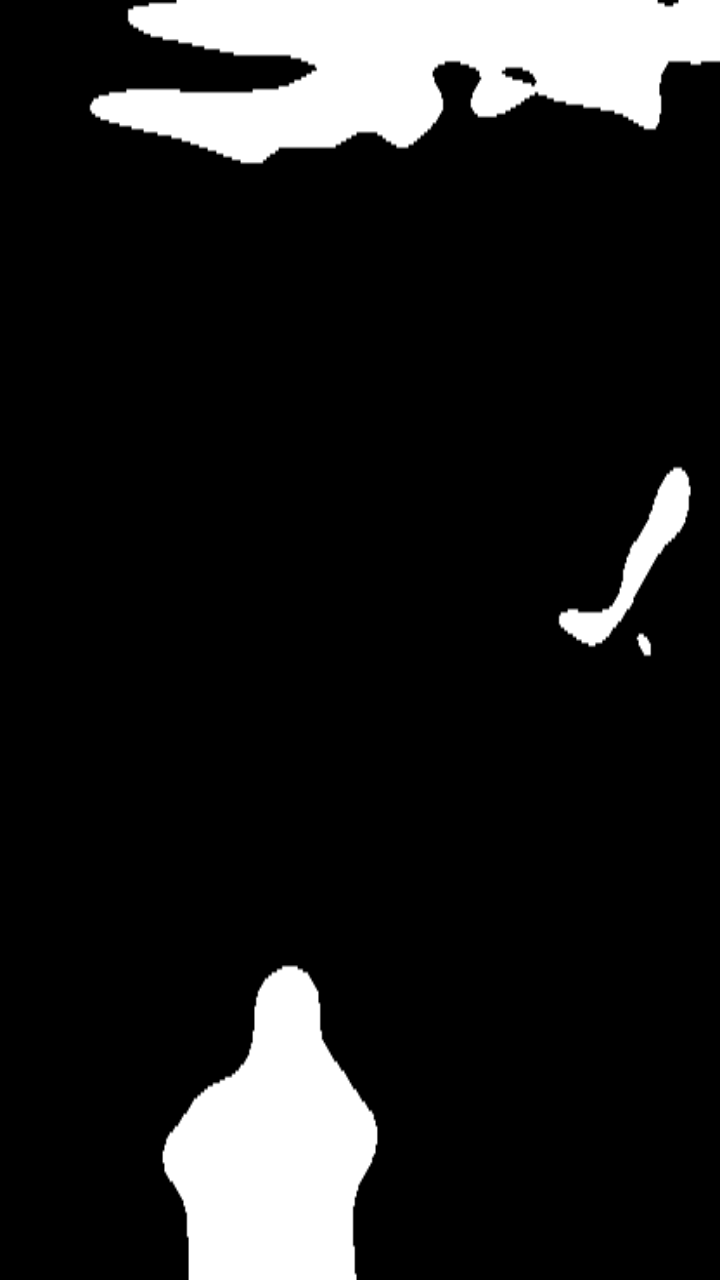}
	\end{subfigure}
	\ \\
	\vspace*{0.5mm}
    \begin{subfigure}{0.085\textwidth} 
		\includegraphics[width=\textwidth]{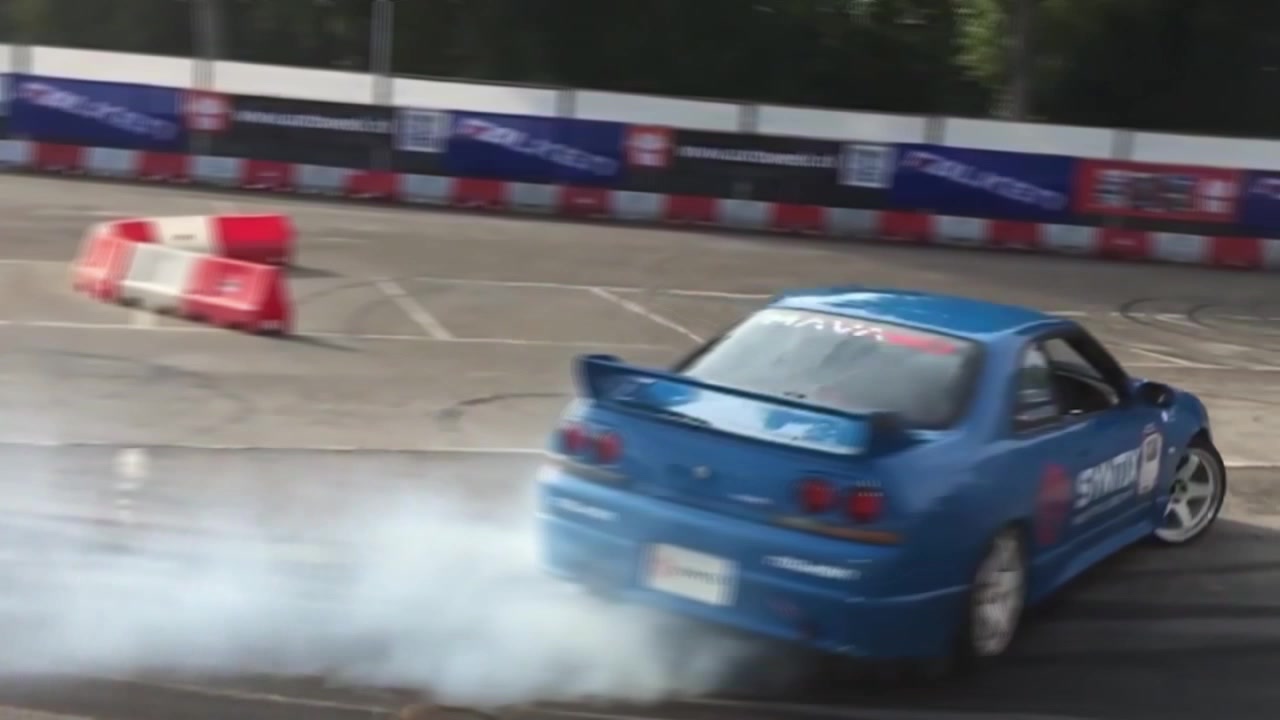}
	\end{subfigure}
	\begin{subfigure}{0.085\textwidth}
		\includegraphics[width=\textwidth]{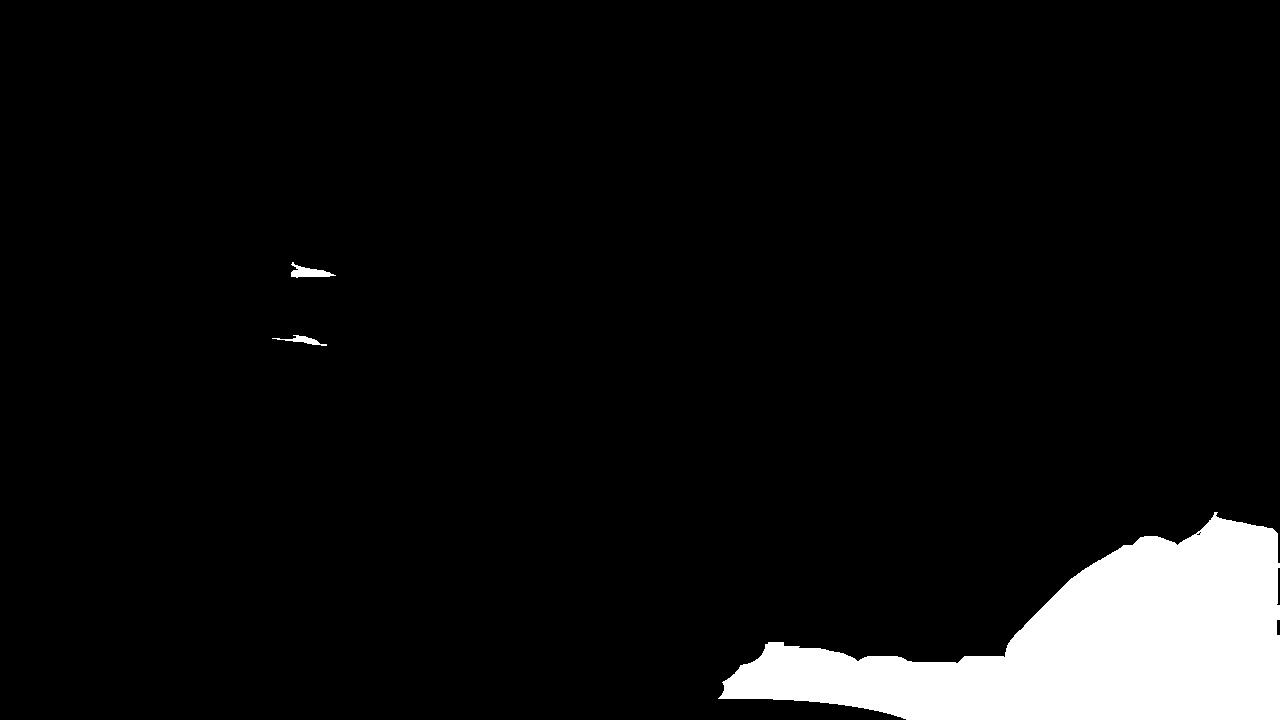}
	\end{subfigure}
	\begin{subfigure}{0.085\textwidth}
		\includegraphics[width=\textwidth]{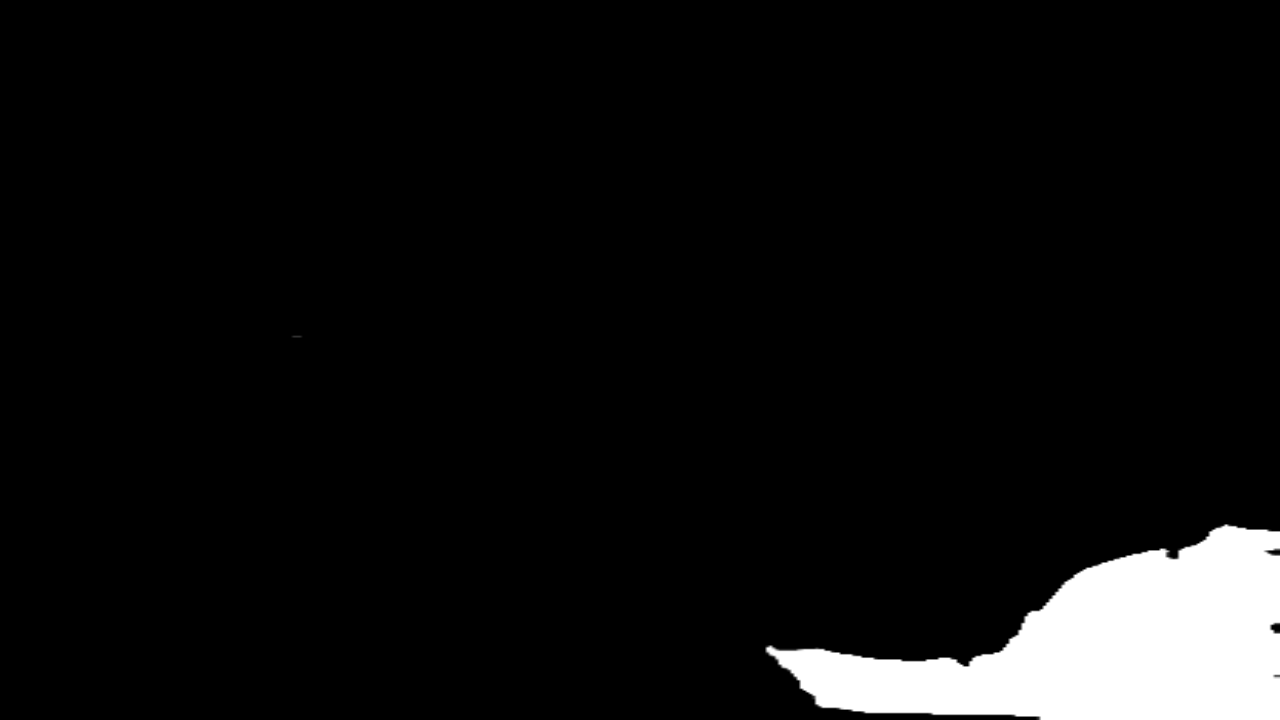}
	\end{subfigure}
	\begin{subfigure}{0.085\textwidth}
		\includegraphics[width=\textwidth]{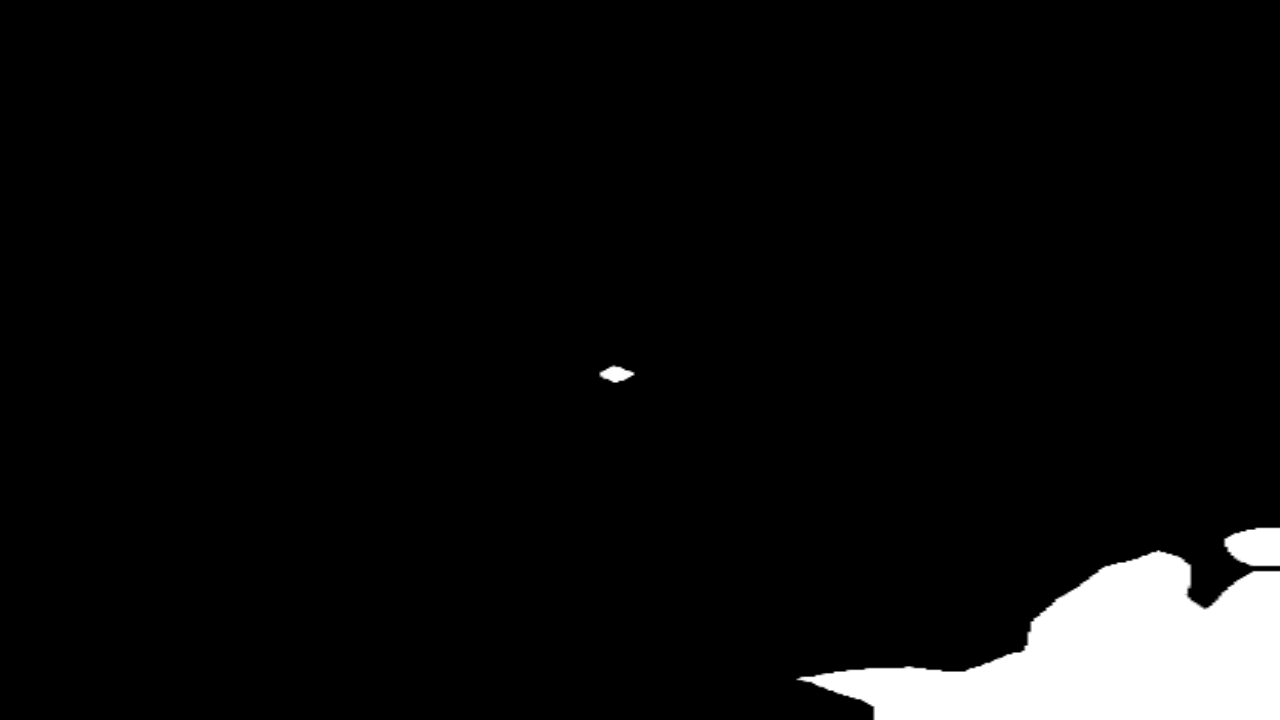}
	\end{subfigure}
	\begin{subfigure}{0.085\textwidth}
		\includegraphics[width=\textwidth]{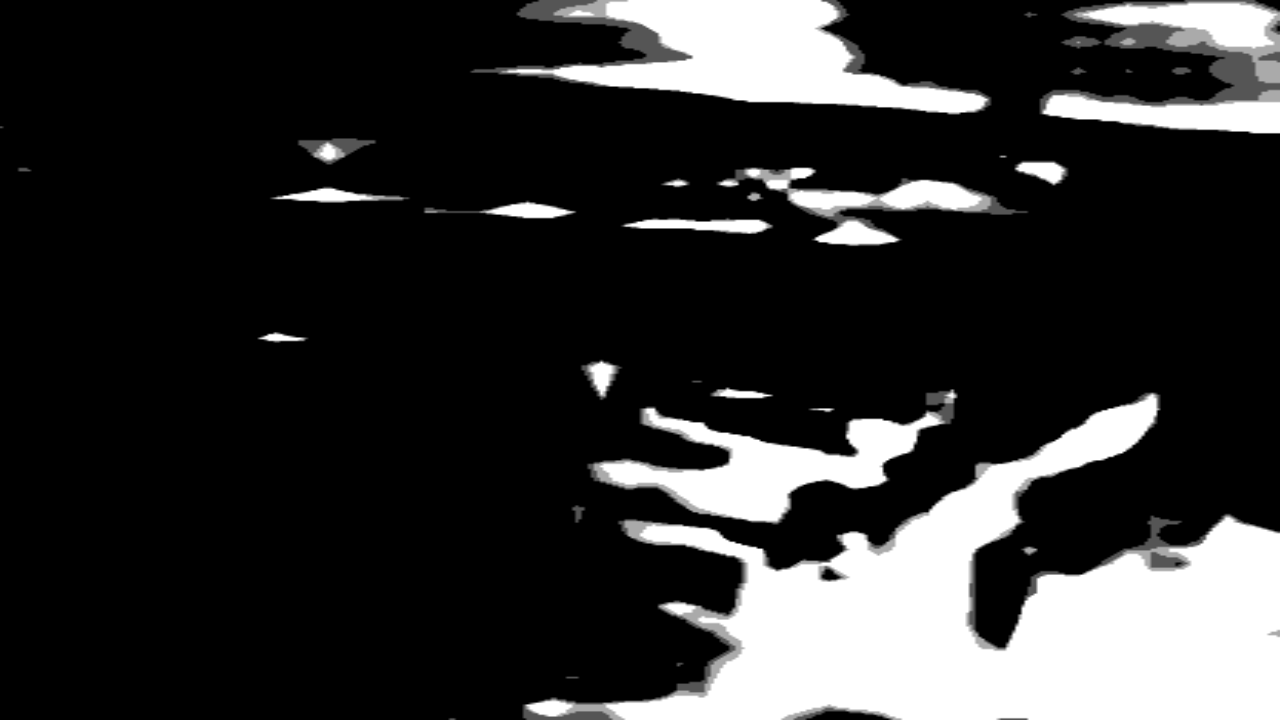}
	\end{subfigure}
	\begin{subfigure}{0.085\textwidth}
		\includegraphics[width=\textwidth]{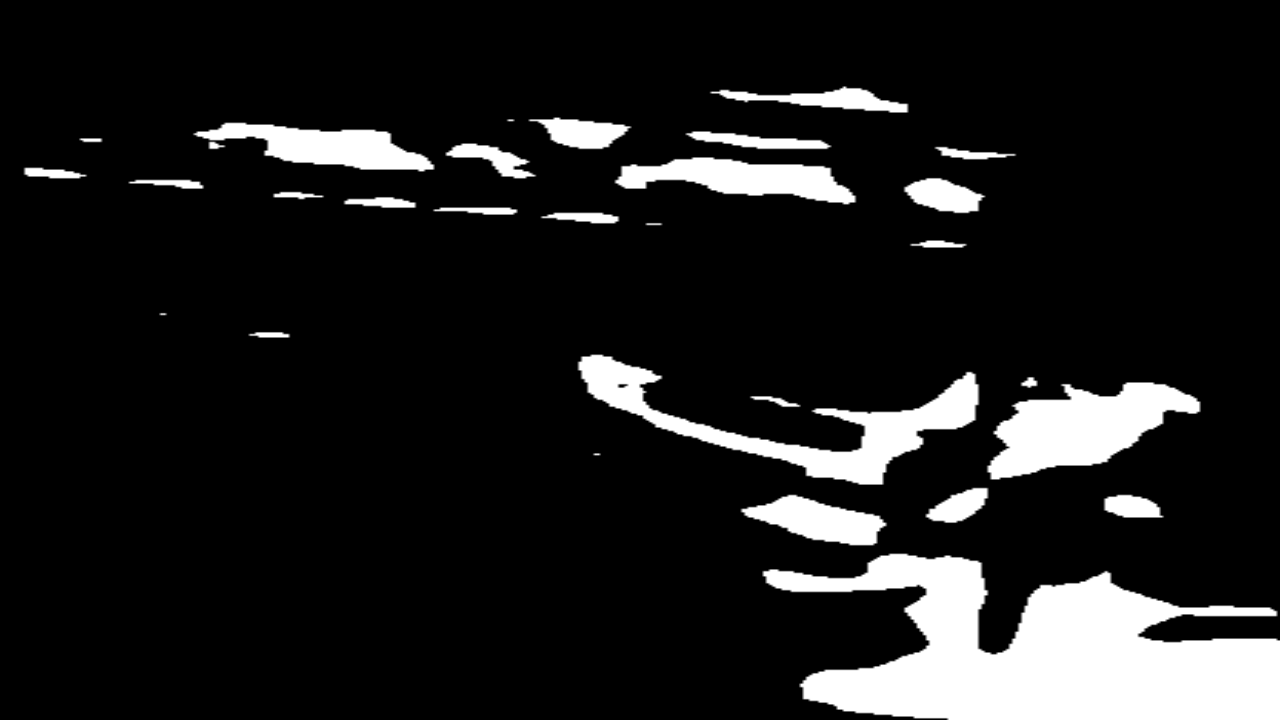}
	\end{subfigure}
	\begin{subfigure}{0.085\textwidth}
		\includegraphics[width=\textwidth]{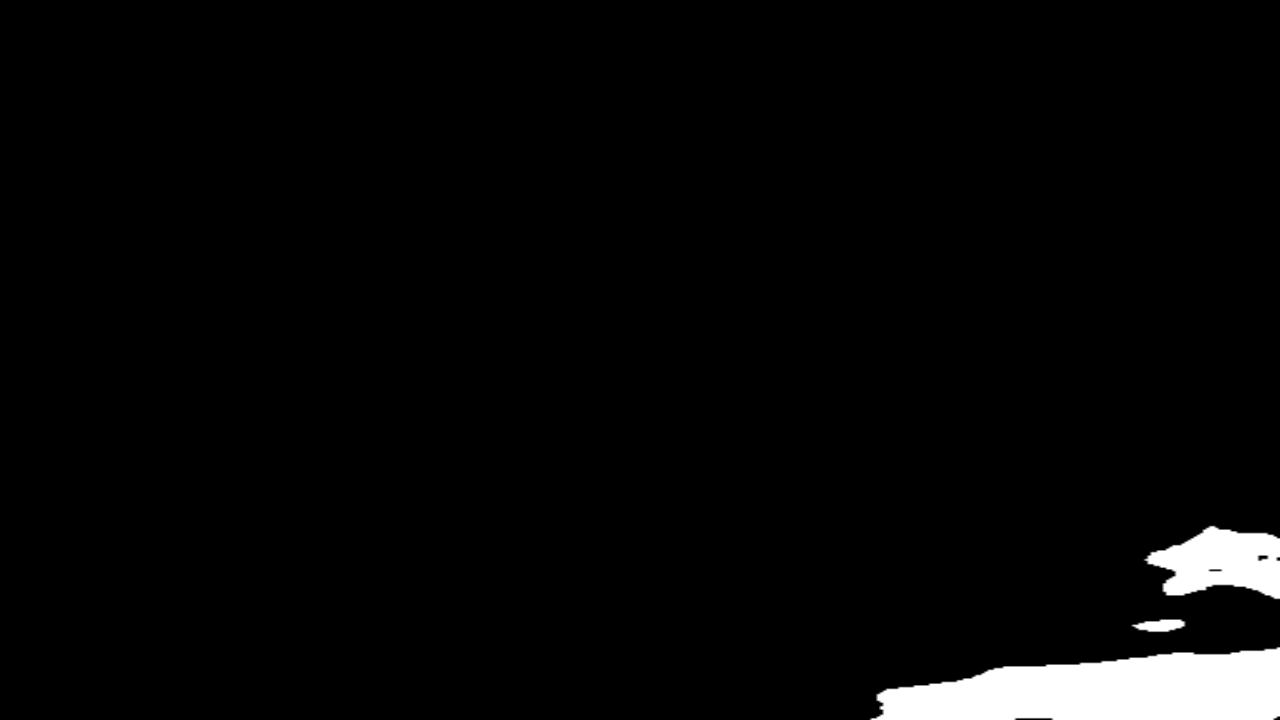}
	\end{subfigure}
	\begin{subfigure}{0.085\textwidth}
		\includegraphics[width=\textwidth]{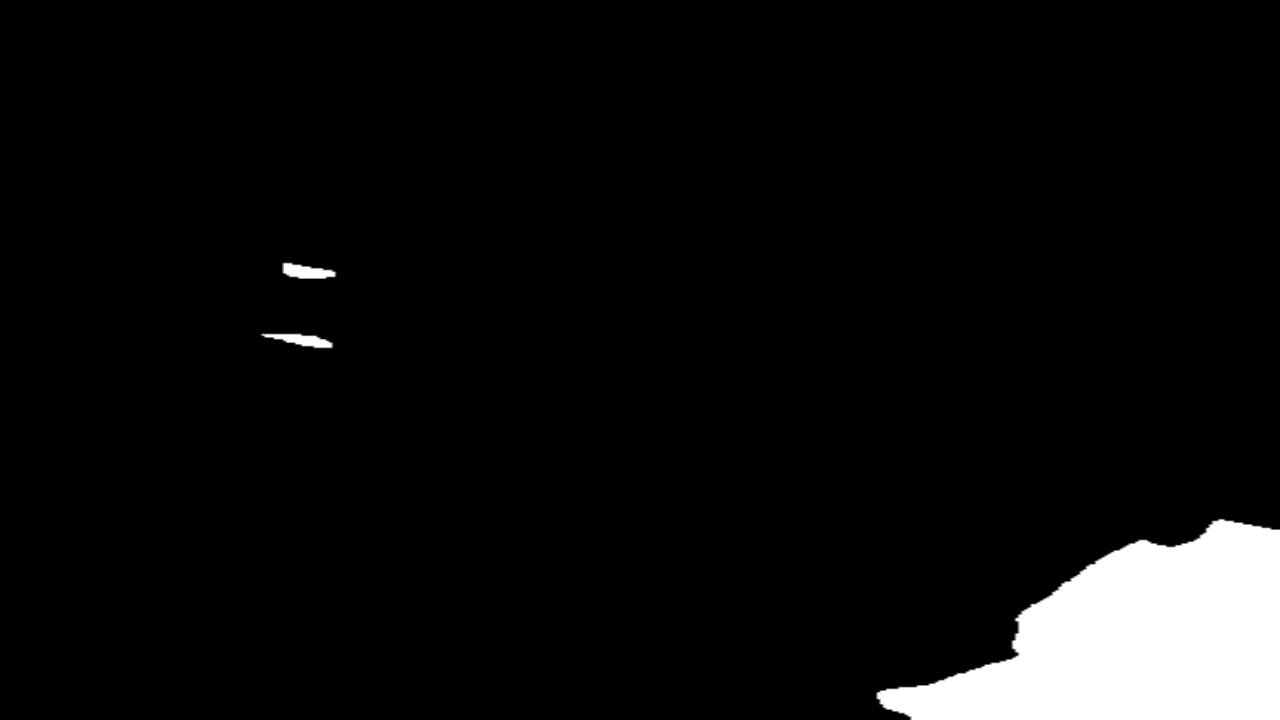}
	\end{subfigure}
	\begin{subfigure}{0.085\textwidth}
		\includegraphics[width=\textwidth]{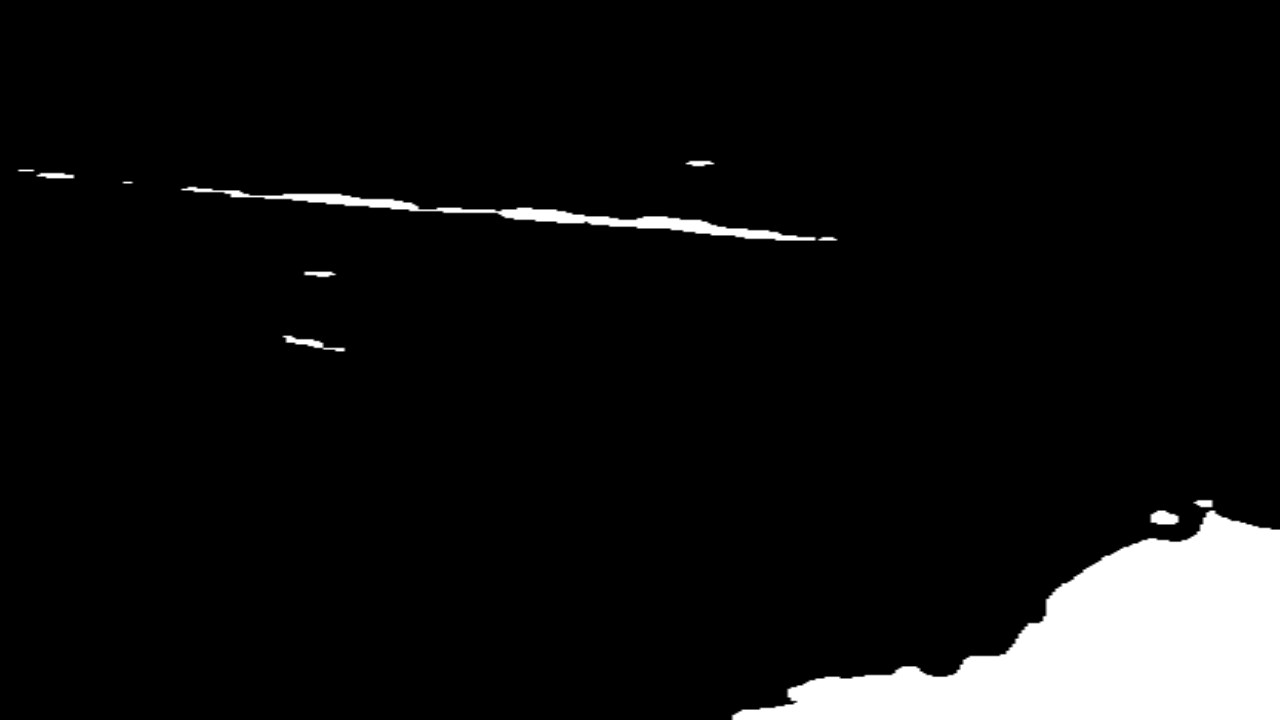}
	\end{subfigure}	
	\begin{subfigure}{0.085\textwidth}
		\includegraphics[width=\textwidth]{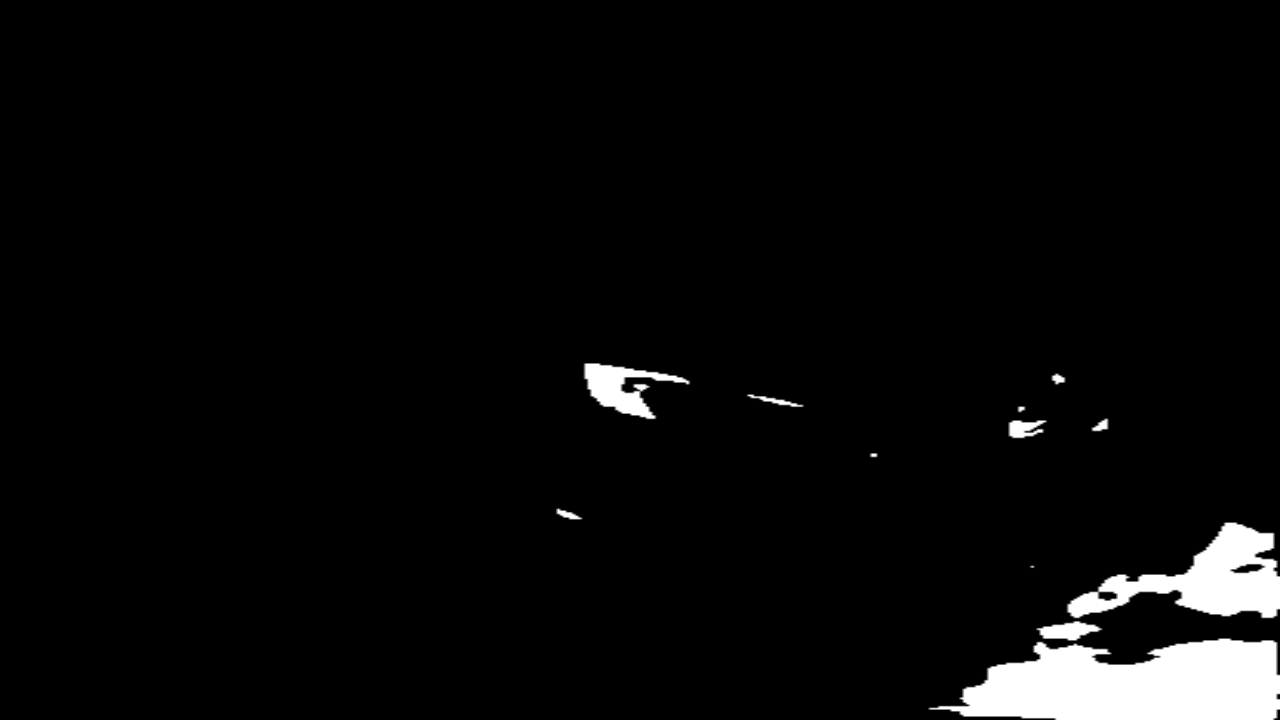}
	\end{subfigure}
	\begin{subfigure}{0.085\textwidth}
		\includegraphics[width=\textwidth]{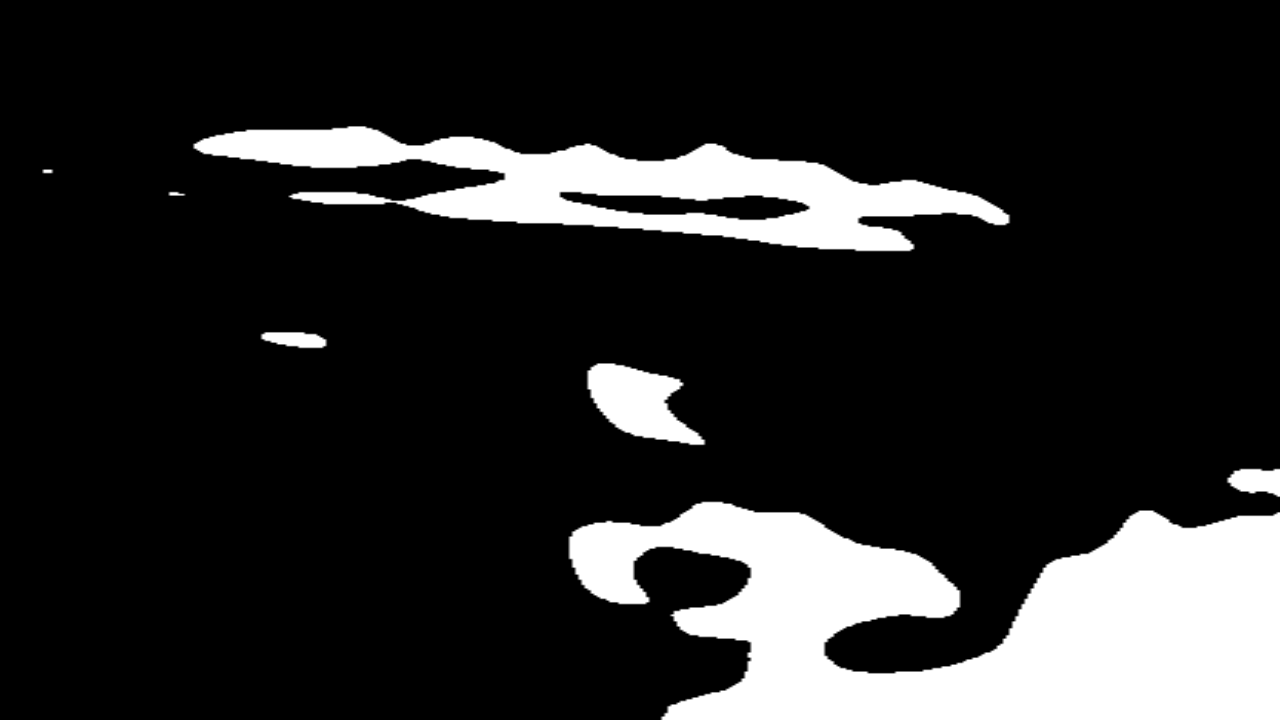}
	\end{subfigure}
	\ \\
	\vspace*{0.5mm}
    \begin{subfigure}{0.085\textwidth} 
		\includegraphics[width=\textwidth]{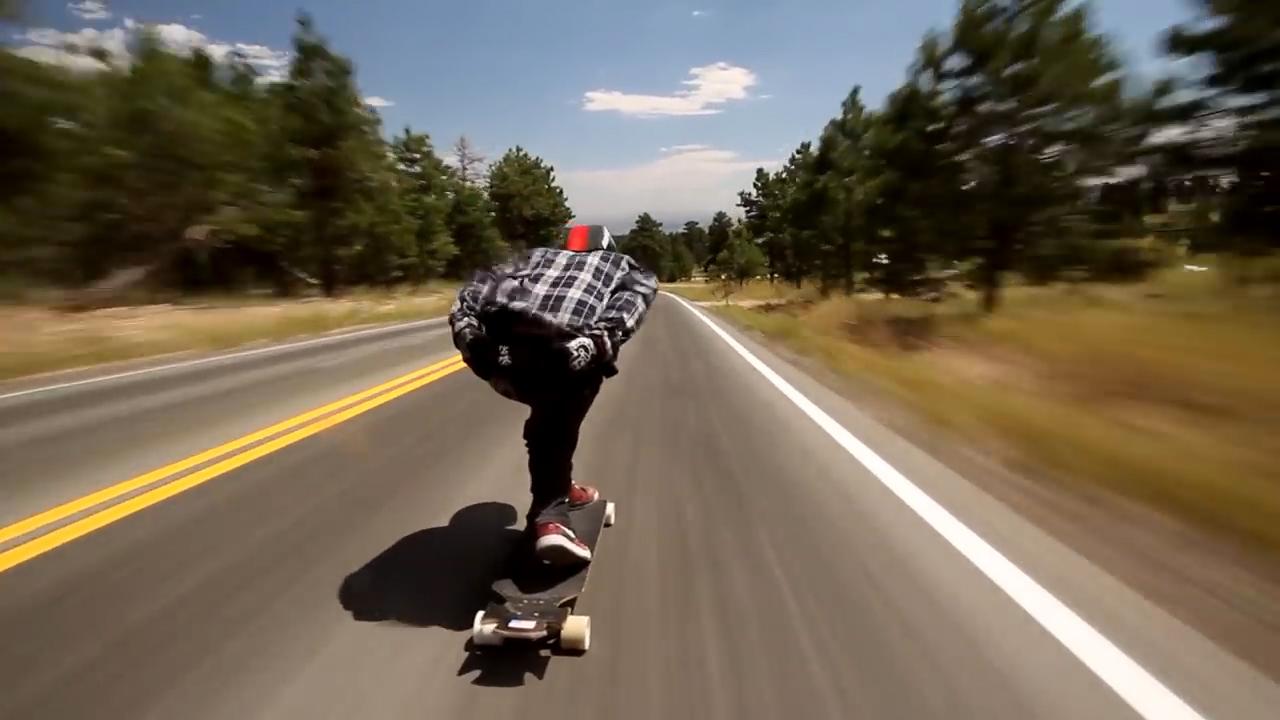}
	\end{subfigure}
	\begin{subfigure}{0.085\textwidth}
		\includegraphics[width=\textwidth]{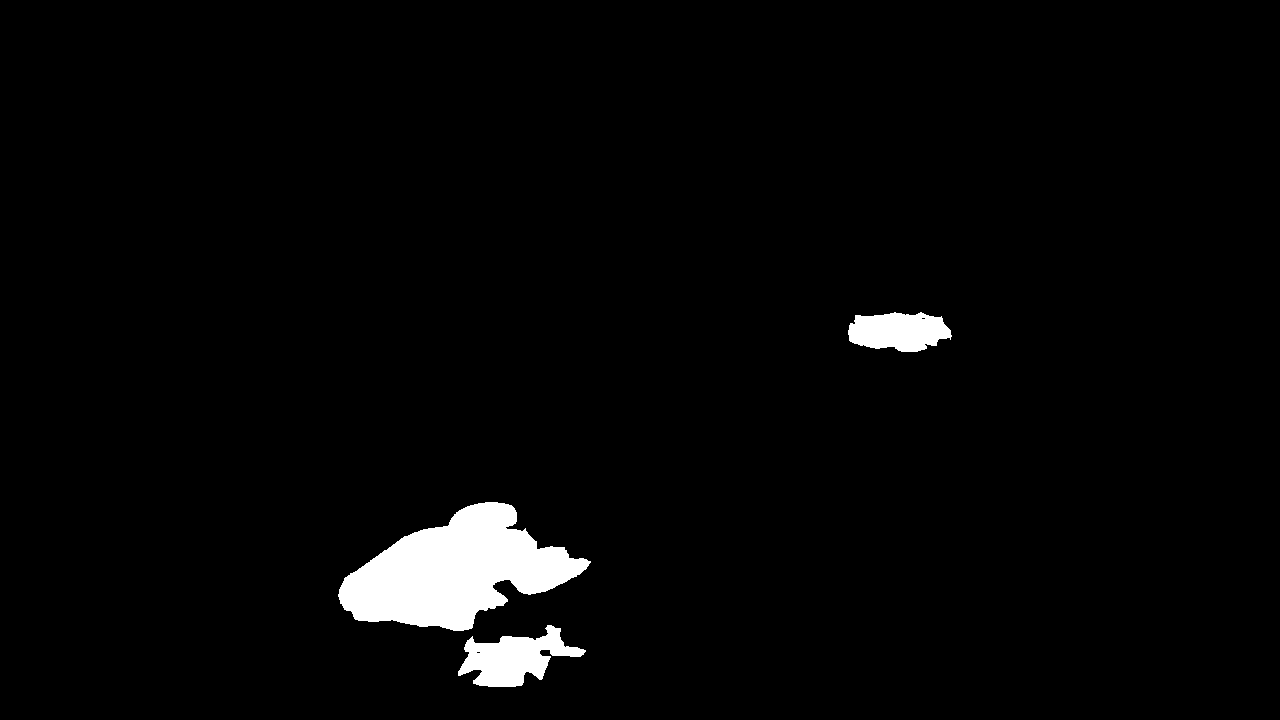}
	\end{subfigure}
	\begin{subfigure}{0.085\textwidth}
		\includegraphics[width=\textwidth]{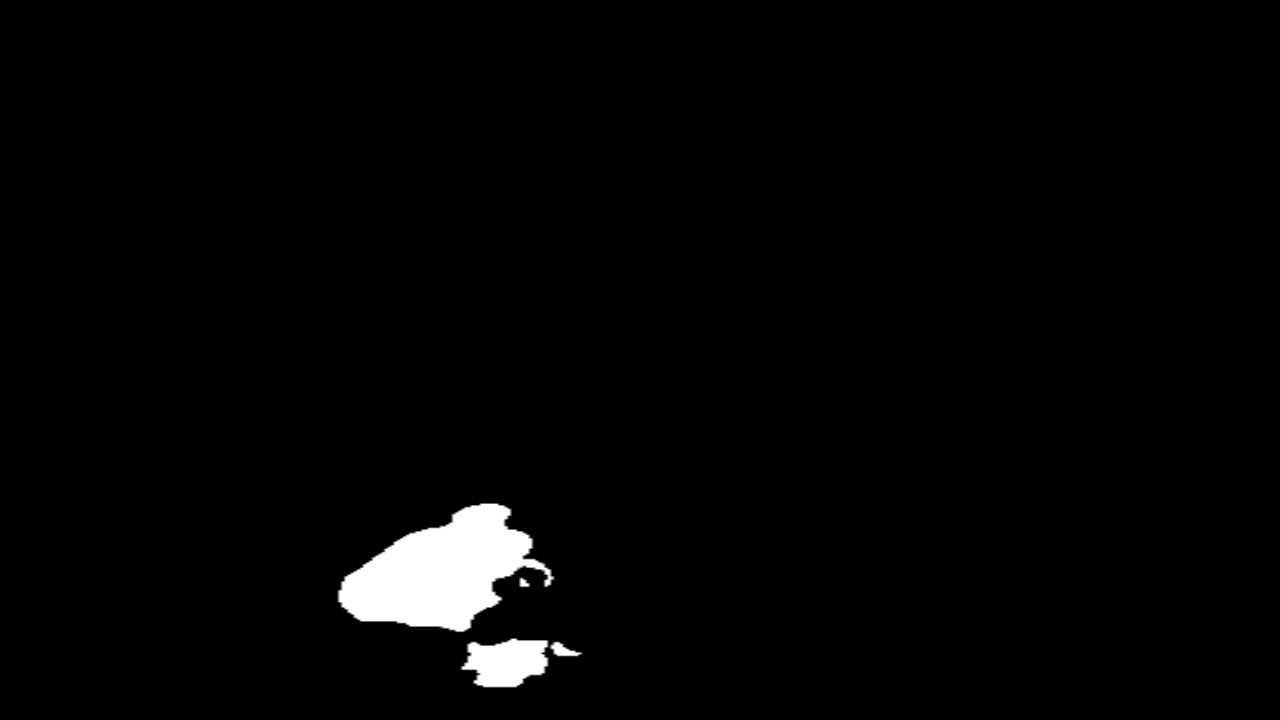}
	\end{subfigure}
	\begin{subfigure}{0.085\textwidth}
		\includegraphics[width=\textwidth]{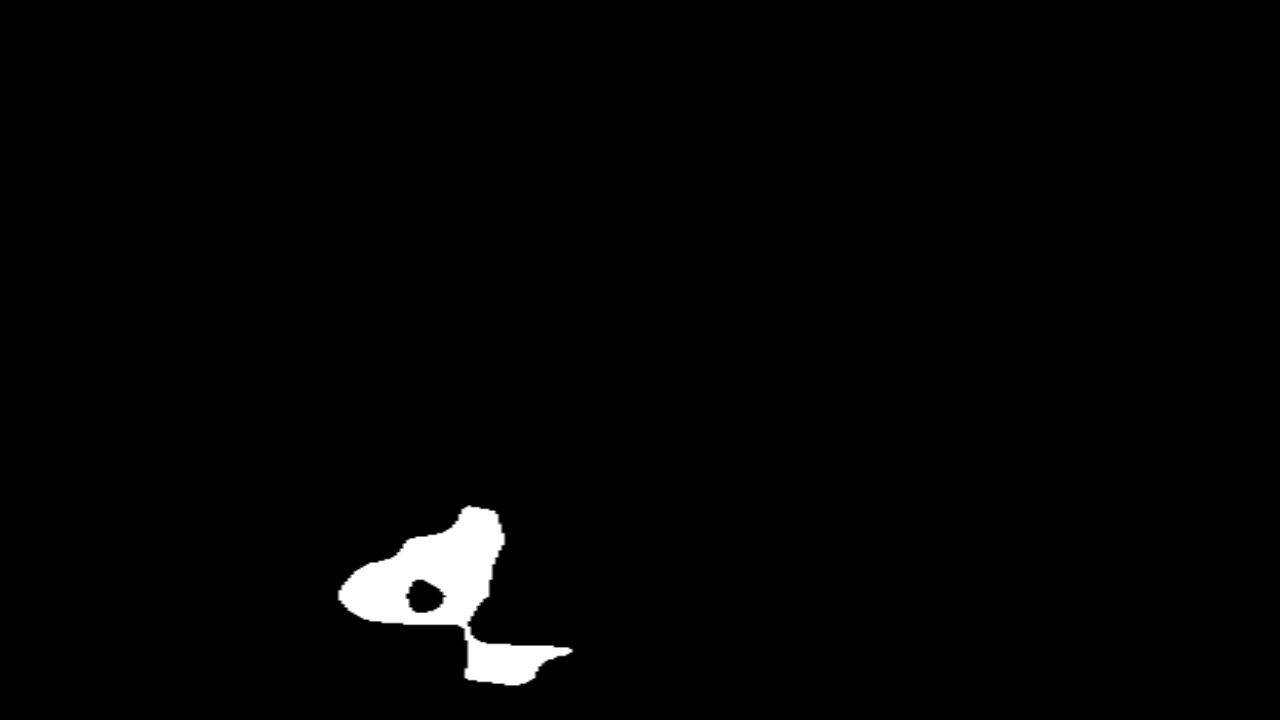}
	\end{subfigure}
	\begin{subfigure}{0.085\textwidth}
		\includegraphics[width=\textwidth]{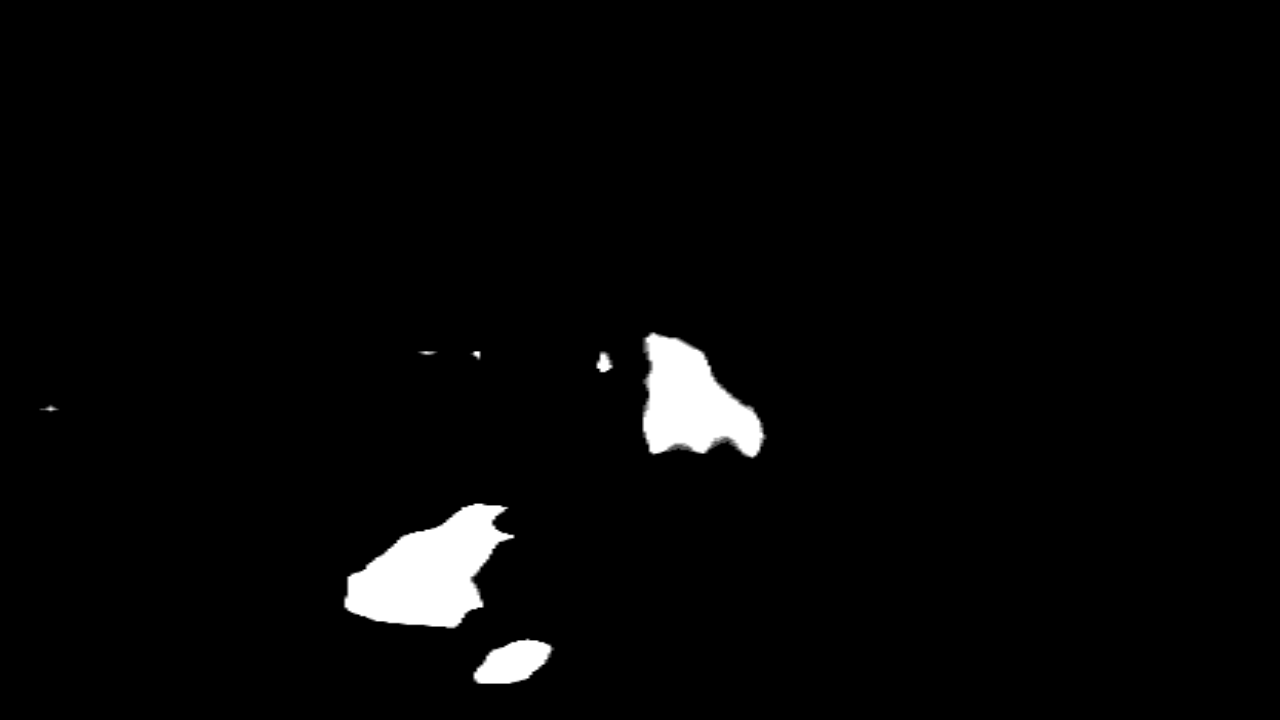}
	\end{subfigure}
	\begin{subfigure}{0.085\textwidth}
		\includegraphics[width=\textwidth]{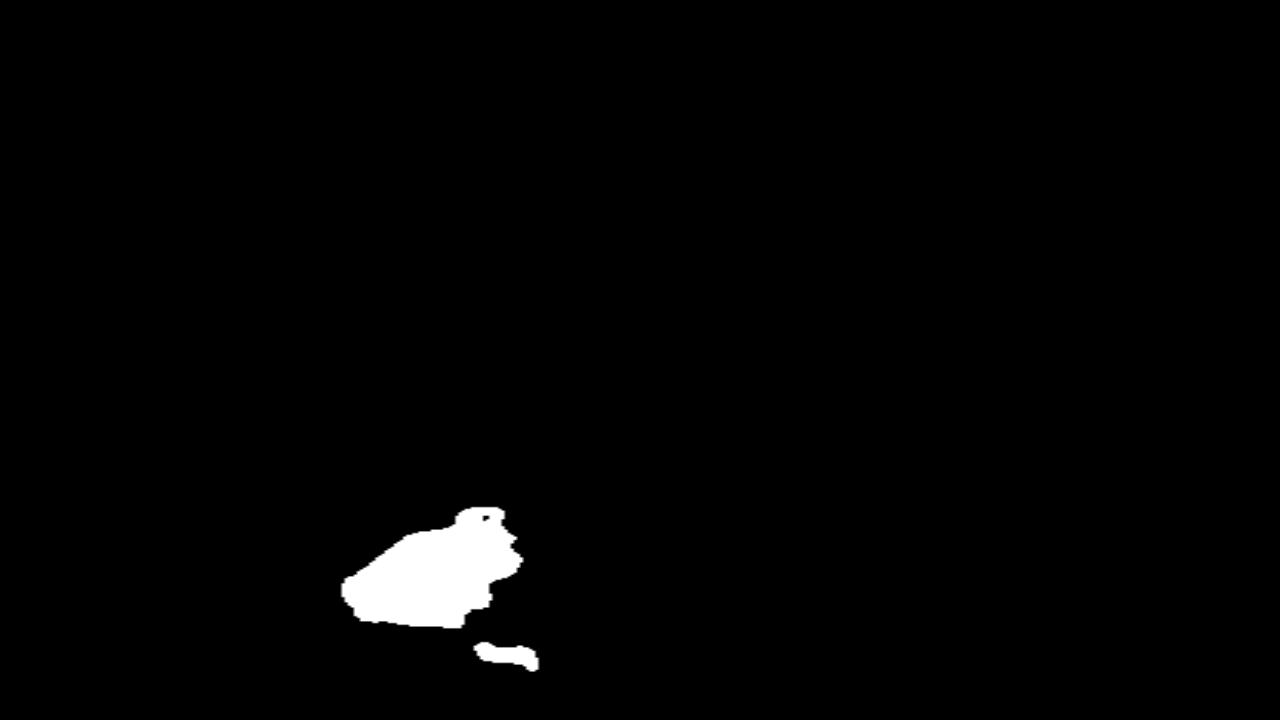}
	\end{subfigure}
	\begin{subfigure}{0.085\textwidth}
		\includegraphics[width=\textwidth]{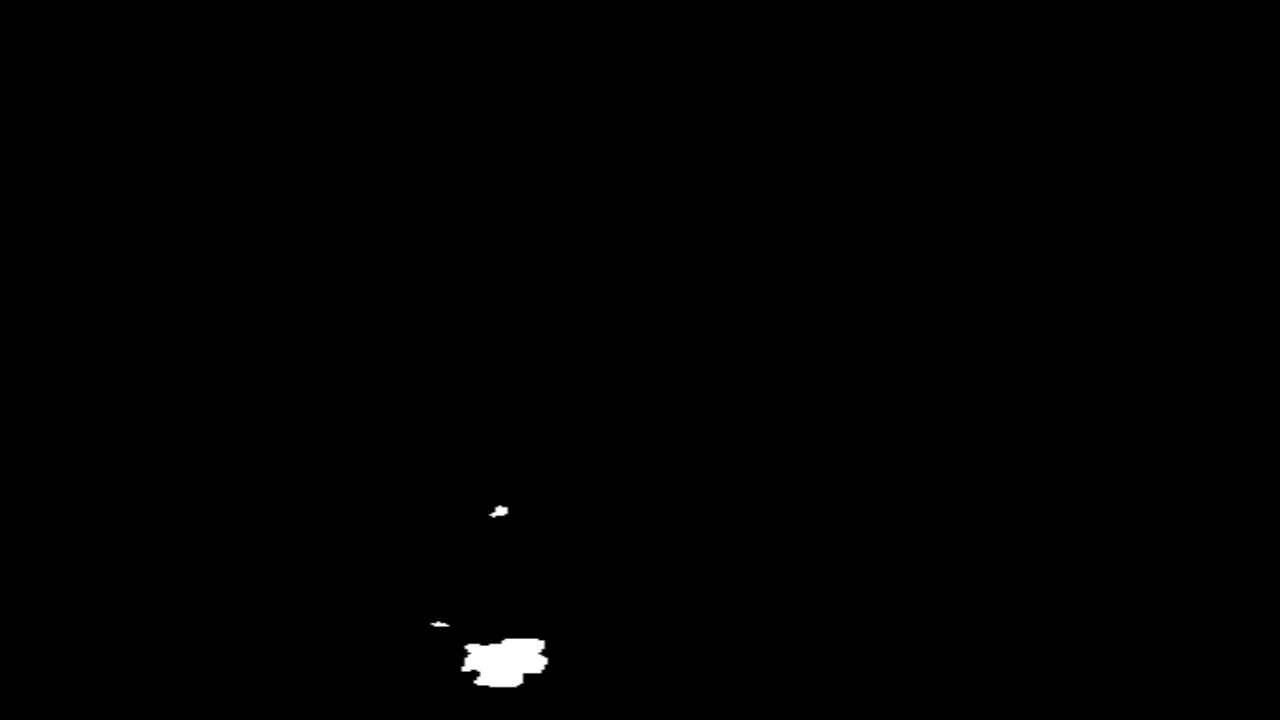}
	\end{subfigure}
	\begin{subfigure}{0.085\textwidth}
		\includegraphics[width=\textwidth]{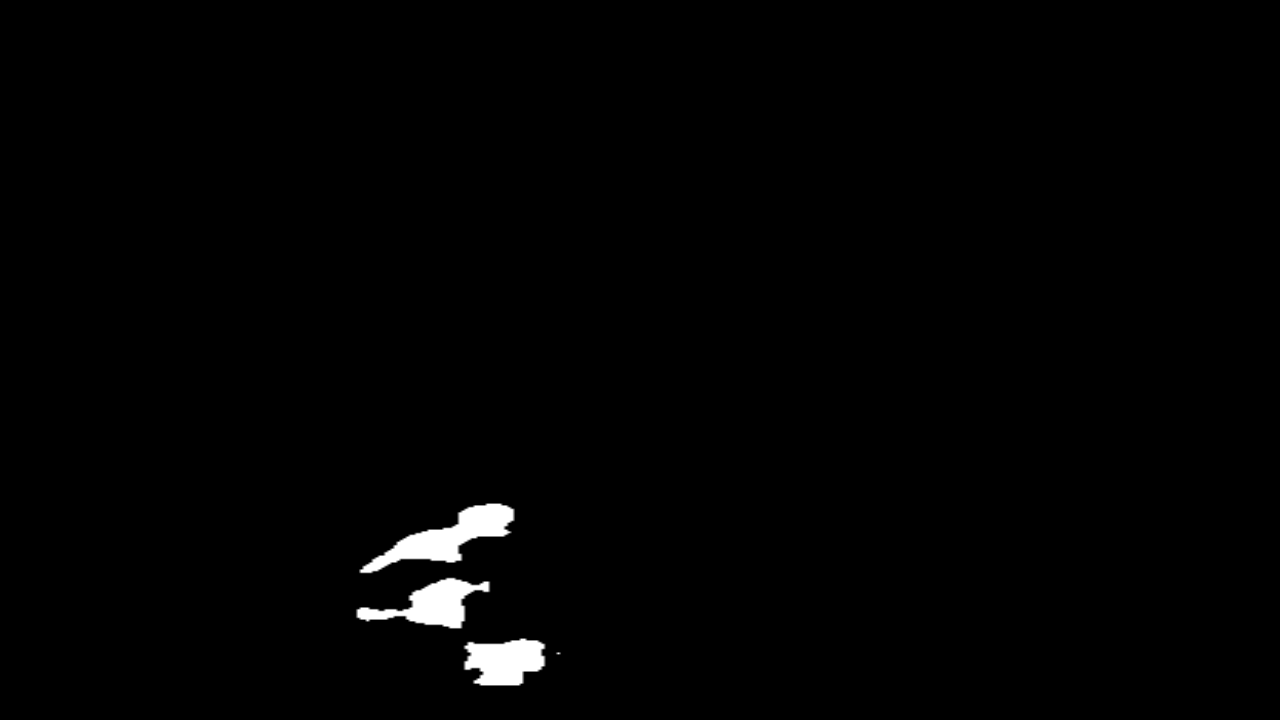}
	\end{subfigure}
	\begin{subfigure}{0.085\textwidth}
		\includegraphics[width=\textwidth]{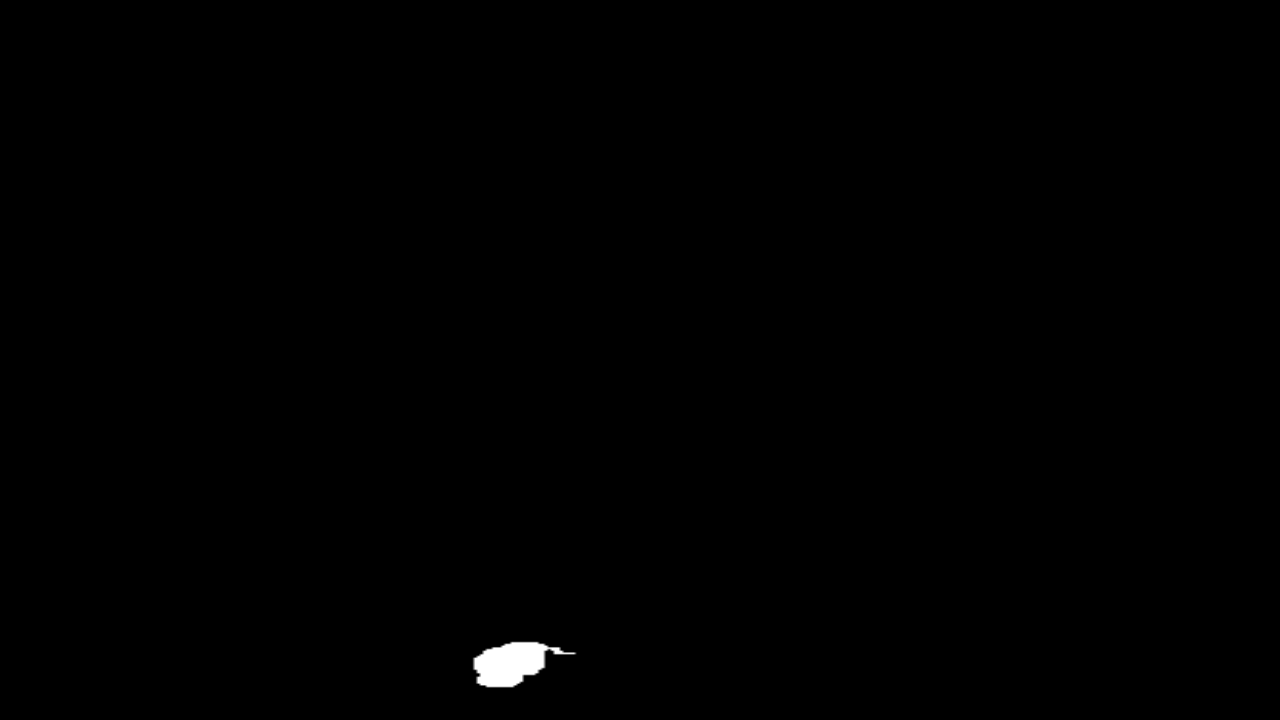}
	\end{subfigure}	
	\begin{subfigure}{0.085\textwidth}
		\includegraphics[width=\textwidth]{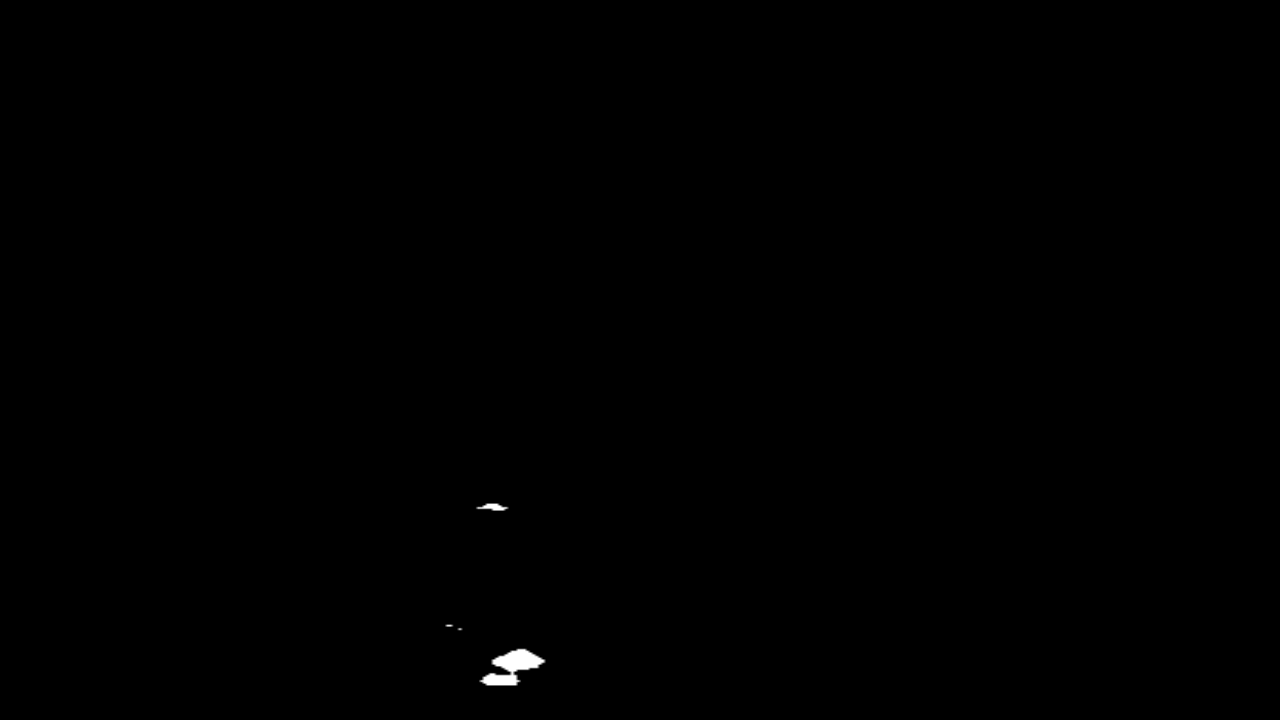}
	\end{subfigure}
	\begin{subfigure}{0.085\textwidth}
		\includegraphics[width=\textwidth]{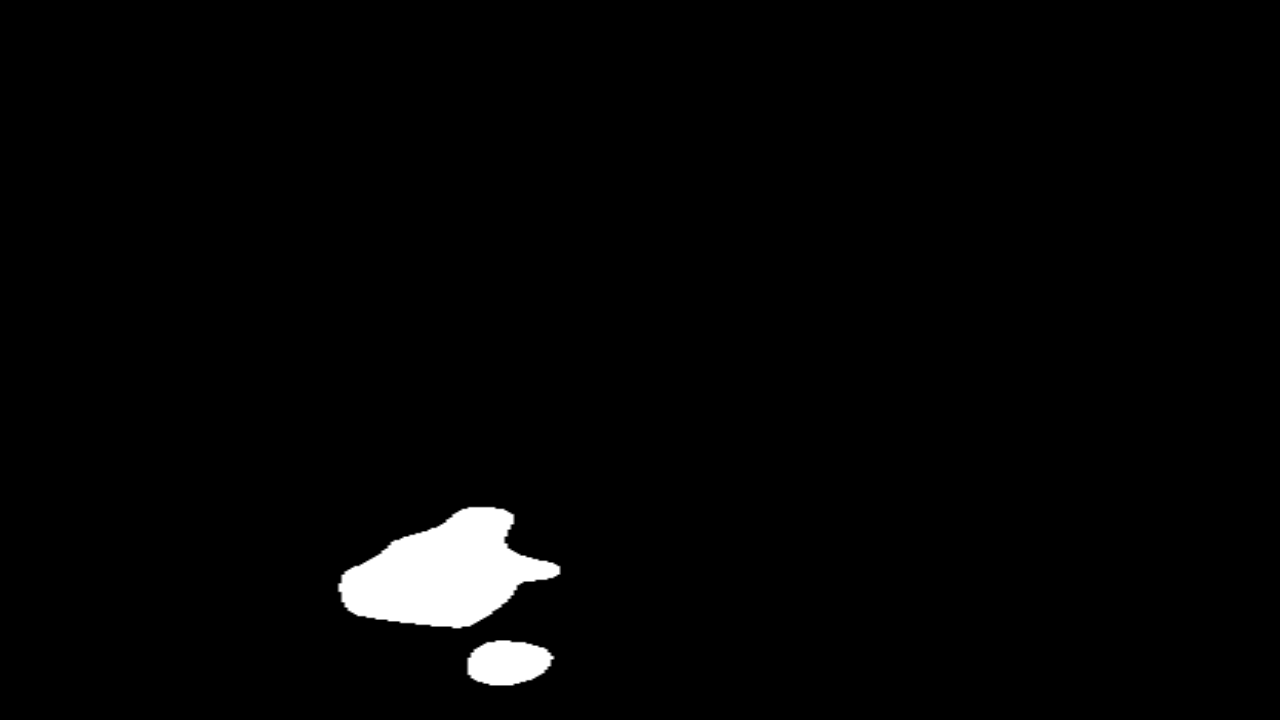}
	\end{subfigure}
	\ \\
	\vspace*{0.5mm}
    \begin{subfigure}{0.085\textwidth} 
		\includegraphics[width=\textwidth]{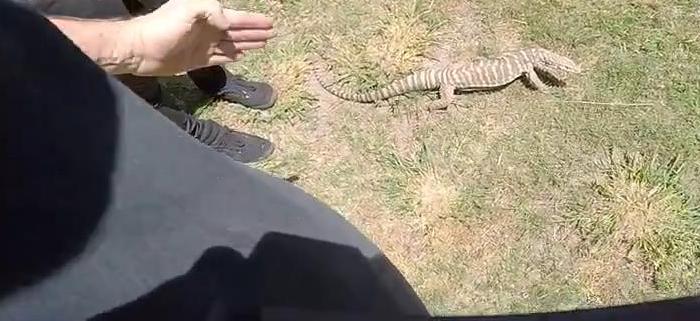}
		\vspace{-5.5mm} \caption*{{\footnotesize Input}}
        \vspace{-2mm} \caption*{\hspace*{-0.7mm}{\footnotesize images}}
	\end{subfigure}
	\begin{subfigure}{0.085\textwidth}
		\includegraphics[width=\textwidth]{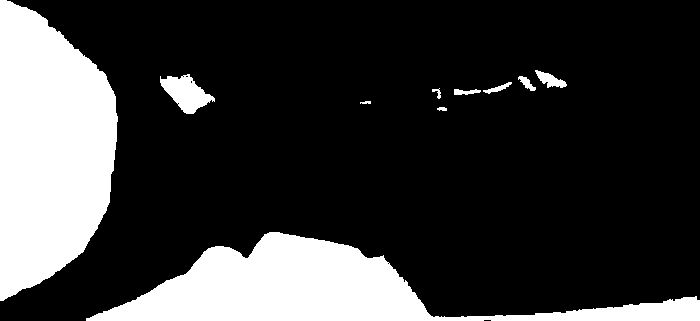}
		\vspace{-5.5mm} \caption*{{\footnotesize Ground}}
        \vspace{-2mm} \caption*{\hspace*{-0.7mm}{\footnotesize truths}}
	\end{subfigure}
	\begin{subfigure}{0.085\textwidth}
		\includegraphics[width=\textwidth]{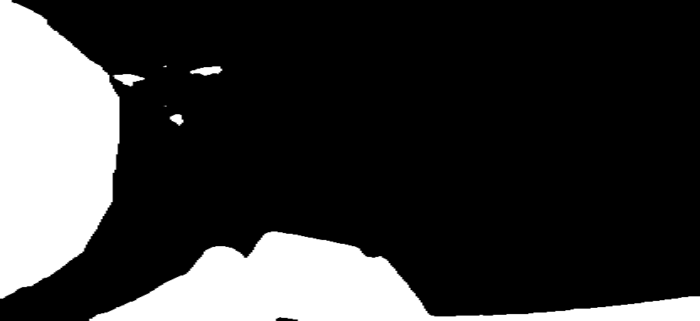}
		\vspace{-5.5mm} \caption*{{\footnotesize TVSD-Net}}
        \vspace{-2mm} \caption*{\hspace*{-0.7mm}{\footnotesize (ours)}}
	\end{subfigure}
	\begin{subfigure}{0.085\textwidth}
		\includegraphics[width=\textwidth]{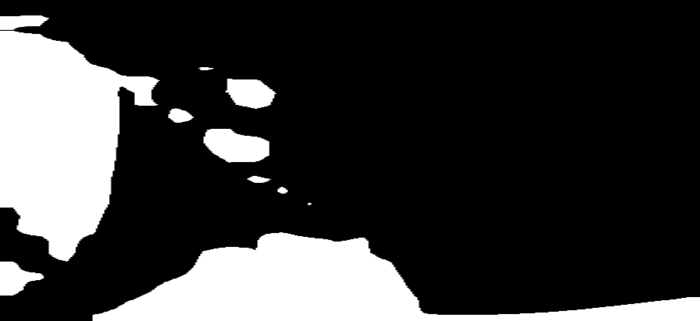}
		\vspace{-5.5mm} \caption*{{\footnotesize COSNet~\cite{lu2019see}}}
        \vspace{-2mm} \caption*{\hspace*{-0.7mm}{}}
	\end{subfigure}
	\begin{subfigure}{0.085\textwidth}
		\includegraphics[width=\textwidth]{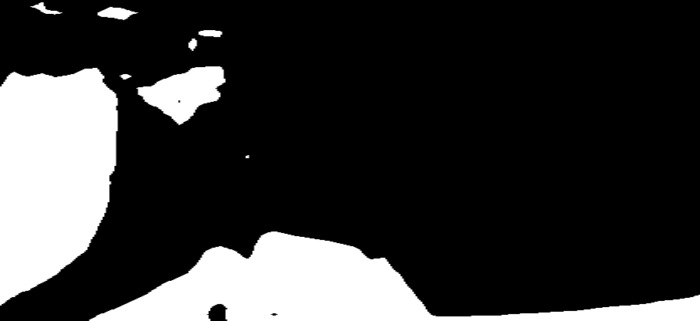}		\vspace{-5.5mm} \caption*{{\footnotesize PDBM~\cite{song2018pyramid}}}
        \vspace{-2mm} \caption*{\hspace*{-0.7mm}{}}
		
	\end{subfigure}
	\begin{subfigure}{0.085\textwidth}
		\includegraphics[width=\textwidth]{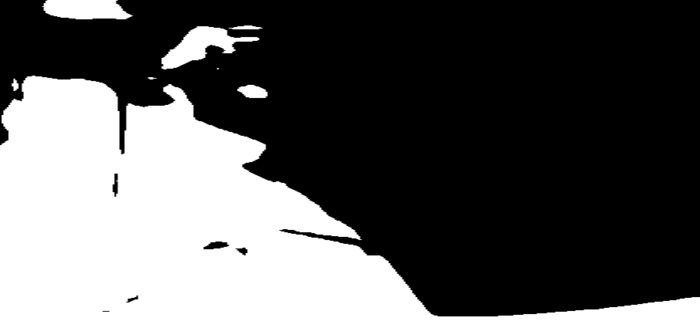}		\vspace{-5.5mm} \caption*{{\footnotesize MGA~\cite{li2019motion}}}
        \vspace{-2mm} \caption*{\hspace*{-0.7mm}{}}
	\end{subfigure}
	\begin{subfigure}{0.085\textwidth}
		\includegraphics[width=\textwidth]{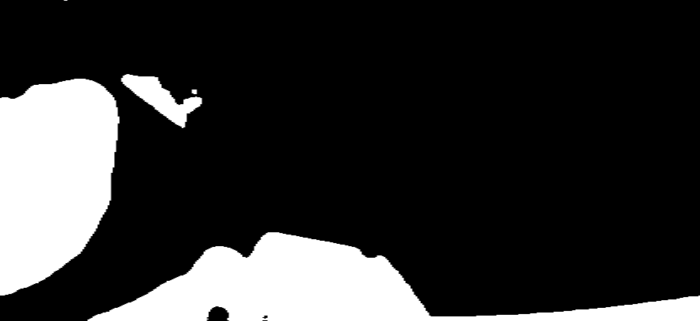}		\vspace{-5.5mm} \caption*{{\footnotesize MTMT~\cite{chen2020multi}}}
        \vspace{-2mm} \caption*{\hspace*{-0.7mm}{}}
	\end{subfigure}
	\begin{subfigure}{0.085\textwidth}
		\includegraphics[width=\textwidth]{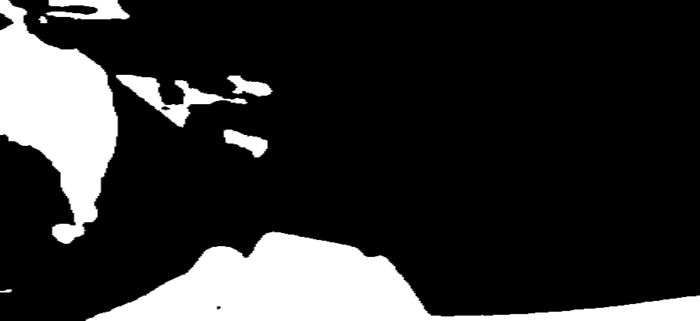}		\vspace{-5.5mm} \caption*{{\footnotesize DSD~\cite{zheng2019distraction}}}
        \vspace{-2mm} \caption*{\hspace*{-0.7mm}{}}
	\end{subfigure}
	\begin{subfigure}{0.085\textwidth}
		\includegraphics[width=\textwidth]{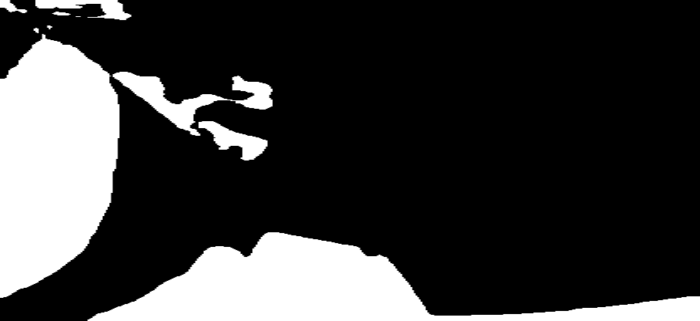}		\vspace{-5.5mm} \caption*{{\footnotesize BDRAR~\cite{zhu2018bidirectional}}}
        \vspace{-2mm} \caption*{\hspace*{-0.7mm}{}}
	\end{subfigure}	
	\begin{subfigure}{0.085\textwidth}
		\includegraphics[width=\textwidth]{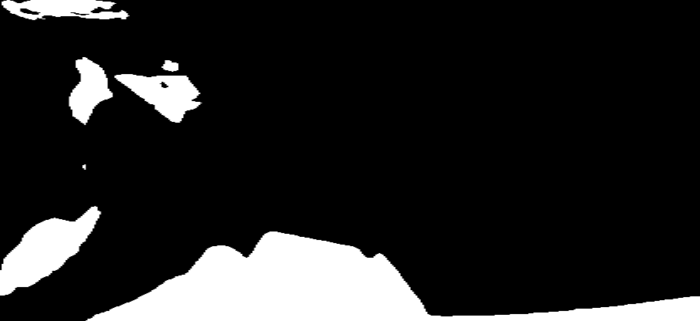}		\vspace{-5.5mm} \caption*{{\footnotesize R$^3$Net~\cite{deng2018r3net}}}
        \vspace{-2mm} \caption*{\hspace*{-0.7mm}{}}
	\end{subfigure}
	\begin{subfigure}{0.085\textwidth}
		\includegraphics[width=\textwidth]{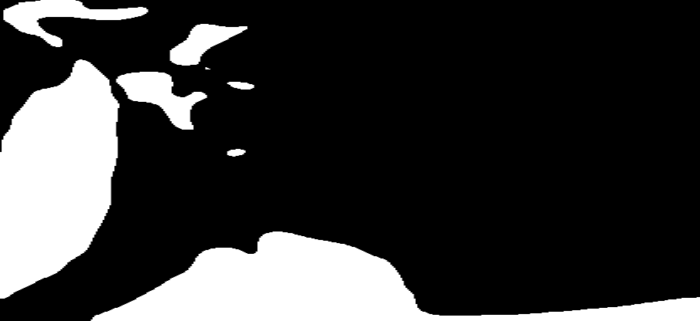}		\vspace{-5.5mm} \caption*{{\footnotesize PSPNet~\cite{Zhao_2017_CVPR}}}
        \vspace{-2mm} \caption*{\hspace*{-0.7mm}{}}
	\end{subfigure}
	
	\ \\
	
    \vspace{-3.5mm}
	\caption{Visual comparison of shadow maps produced by our method and other methods. }
	\label{fig:comparison_SOTA}
    \vspace{-4.5mm}
\end{figure*}

\begin{table}[t]
\begin{center}
  \caption{Comparing our network (TVSD-Net) against the state-of-the-art methods. }
  \vskip -5pt
  \label{table:state-of-the-art}
  \resizebox{0.45\textwidth}{!}{%
    \begin{tabular}{c|c|c|c|c|c}
        \toprule[1pt]
        Method & Year & MAE $\downarrow$ & $\mathrm{F}_{\beta}$ $\uparrow$ & IoU $\uparrow$ & BER$\downarrow$ \\
        \hline
        \hline
        BDRAR & 2018 & 0.050 & 0.695 & 0.484 & 21.29 \\
        \hline
        DSD & 2019 & 0.044 & 0.702 & 0.518 & 19.88 \\
        \hline
        MTMT & 2020 & 0.043 & 0.729 & 0.517 & 20.28 \\
        \hline
        \hline
        FPN & 2017 & 0.044 & 0.707 & 0.512 & 19.49 \\
        \hline
        PSPNet & 2017 & 0.051 & 0.642 & 0.476 & 19.75 \\
        \hline
        \hline
        DSS & 2017 & 0.045 & 0.696 & 0.502 & 19.77 \\
        \hline
        $\mathrm{R}^3$Net & 2018 & 0.043 & 0.710 & 0.502 & 20.40 \\
        \hline
        \hline
        PDBM & 2018 & 0.066 & 0.623 & 0.466 & 19.73 \\
        \hline
        COSNet & 2019 & 0.040 & 0.705 & 0.514 & 20.50 \\
        \hline
        MGA & 2019 & 0.067 & 0.601 & 0.399 & 25.77 \\
        \hline
        FEELVOS & 2019 & 0.043 & 0.710 & 0.512 & 19.76\\
        \hline
        STM & 2019 & 0.068 & 0.597 & 0.408 & 25.69\\
        \hline
        \hline
        TVSD-Net (Ours) & - & \textbf{0.033} & \textbf{0.757} & \textbf{0.567} & \textbf{17.70} \\
        \bottomrule[1pt]
    \end{tabular}
    }
    \vspace{-5mm}
  \end{center}
\end{table}

\begin{table}[!t]
	\begin{center}
		\caption{Ablation analysis. Here, “Co-att” denotes the original Co-attention module in \cite{lu2019see}; “DGM” denotes our proposed Dual-Gated Mechanism; “T-module” denotes the Triple-cooperative module.}
        \vspace{-2mm}	
		\label{table:ablation}
		\resizebox{\columnwidth}{!}{
		\begin{tabular}{c|c|c|c|c|c|c|c}
		    \toprule[1pt]
			Network & Co-att& DGM & T-m & MAE $\downarrow$ & $\mathrm{F}_{\beta}$ $\uparrow$ & IoU $\uparrow$ & BER$\downarrow$  \\
			\hline
			\hline
			basic & $\times$ & $\times$ & $\times$ & 0.042 & 0.743 & 0.538 & 19.17 \\
			\hline
			basic+Co-att & $\checkmark$ & $\times$ & $\times$ & 0.041 & 0.739 & 0.545 & 18.53 \\
			\hline			
			basic+T-module & $\times$ & $\times$ & $\checkmark$ & 0.039 & 0.739 & 0.549 & 18.57 \\
			\hline
            ours-w/o-T-module & $\checkmark$ & $\checkmark$ & $\times$ & 0.038 & 0.744 & 0.551 & 18.72 \\
            \hline
			ours-w/o-DGM & $\checkmark$& $\times$ & $\checkmark$ & 0.038 & 0.756 & 0.540 & 19.55 \\
			\hline
			our method & $\times$ & $\checkmark$ & $\checkmark$ & \textbf{0.033} & \textbf{0.757} & \textbf{0.567} & \textbf{17.70} \\
			\bottomrule[1pt]
		\end{tabular}
		}
	\end{center}
\vspace{-5mm}	
\end{table}


\subsection{Comparison to the State-of-the-arts}

Table~\ref{table:state-of-the-art} shows the performances on our ViSha dataset, where COSNet has the best $\mathrm{MAE}$ score of $0.040$; MTMT has the best $\mathrm{F}_{\beta}$ score of $0.729$; DSD has the best $\mathrm{IoU}$ score of $0.518$; PDBM has the best $\mathrm{BER}$ score of $19.73$.
Compared to the best-performing existing methods, our method obtains improvements with large margins, with $\mathrm{MAE}$ improvement of 17.50\%, $\mathrm{F}_{\beta}$ improvement of 3.84\%, $\mathrm{IoU}$ improvement of 9.46\%, and $\mathrm{BER}$ improvement of 10.29\% on our dataset, respectively. That shows the superiority of our method on video shadow detection.

Figure~\ref{fig:comparison_SOTA} visually compares the video shadow detection maps produced by our method and the state-of-the-arts.
From the results, we can see that our TVSD-Net (3rd column of Figure~\ref{fig:comparison_SOTA}) can more accurately identify shadow pixels than compared methods. 
It effectively locates different shadows under various backgrounds, and successfully discriminates true shadows from those non-shadow dark regions. 
For example, in the 1st row, most compared methods regard both the bottle and its shadow as the shadow regions, while our TVSD-Net can discriminate them successfully. A similar situation is also reflected in the 2nd row, our TVSD-Net can better capture the distinction between the soccer player and his shadow.

\subsection{Ablation Study} 

We perform ablation study experiments to verify the performance of Dual-gated co-attention module and Triple-cooperative module in TVSD-Net.
Here, we consider four baseline networks.
The first baseline network (denoted as ``basic'') is constructed by removing the co-att, DGM, and T module from the TVSD-Net. 
The second (denoted as ``basic+co-att'') is to add the original co-attention module~\cite{lu2019see} into ``basic''.
The third baseline (denoted as ``basic+T-module'') is to add our T-module into ``basic''. 
The fourth baseline (denoted as ``ours-w/o-T-module'') removes T-module from our network. 
The fifth baseline (denoted as ``ours-w/o-DGM'') removes DGM from our network. 

Table~\ref{table:ablation} summarizes the $\mathrm{BER}$ values of our network and seven baseline networks on the ViSha dataset.
From the results, we have the following observations:
(i) ``basic+co-att'' have the superior performance of four evaluation metrics over ``basic'', which means that learning intra-video coherent information can provide helpful information for video shadow detection.
(ii) ``basic+T-module'' has smaller $\mathrm{BER}$ and $\mathrm{MAE}$ scores and larger $\mathrm{F}_{\beta}$  and $\mathrm{IoU}$ scores than ``basic'', demonstrating that the additional auxiliary loss from the T module incurs detection improvement.
(iii) ``ours-w/o-T-module'' can more accurately detect shadow pixels than ``basic+co-att'' due to its better results of $\mathrm{F}_{\beta}$, $\mathrm{IoU}$, $\mathrm{BER}$, and $\mathrm{MAE}$.
It indicates dual gated module helps to increase the co-attention confidences than the original co-attention module~\cite{lu2019see}.
(iv) our TVSD-Net has better metric results than ``ours-w/o-T-module'' and ``ours-w/o-DGM'', showing that combining the two modules achieves a higher video shadow detection accuracy.
%


\section{Conclusion}
\label{sec:conclusion}

This paper presents a novel network for video shadow detection. 
One of our key contributions is to first collect a learning-oriented video shadow detection (ViSha) dataset, which contains $120$ videos with $11,685$ frames covering various objects and scenes, with pixel-level shadow annotations.
The second contribution is the development of a novel network for video shadow detection, by learning intra-video and inter-video discriminative properties of shadows.
Experimental results on the collected dataset demonstrated that our method consistently outperforms 12 state-of-the-art methods by a large margin.
To the best of our knowledge, this work is the first annotated dataset for video shadow detection, and our ViSha dataset can facilitate further research in video shadow detection. 

\vspace{2mm}
\noindent\textbf{Acknowledgments:} The work is supported by the National Natural Science Foundation of China (Grant No. 61902275, 61572354).

{\small
\bibliographystyle{ieee_fullname}
\bibliography{egbib}
}

\end{document}